\documentclass{article}

\usepackage{arxiv}

\usepackage{microtype}
\usepackage{graphicx}
\usepackage{subfigure}

\usepackage{booktabs} 
\usepackage[normalem]{ulem}
\useunder{\uline}{\ul}{}
\usepackage{multicol}
\usepackage[table,xcdraw]{xcolor}

\usepackage{latexsym}
\usepackage{subcaption}

\usepackage[utf8]{inputenc} 
\usepackage[T1]{fontenc}    
\usepackage{hyperref}       
\usepackage{url}            
\usepackage{booktabs}       
\usepackage{amsfonts}       
\usepackage{algorithm}
\usepackage{algpseudocode}
\usepackage{amsmath}
\usepackage{amssymb}
\usepackage{mathtools}
\usepackage{amsthm}
\usepackage{mwe}

\usepackage{nicefrac}       
\usepackage{microtype}      
\usepackage{cleveref}       
\usepackage{lipsum}         
\usepackage{graphicx}
\usepackage{natbib}
\usepackage{doi}

\DeclareMathOperator{\argmin}{arg\,min}

\title{Time-Series Classification for  \\
           Dynamic Strategies in Multi-Step Forecasting}

\newif\ifuniqueAffiliation
\uniqueAffiliationtrue

\ifuniqueAffiliation 
\author{ \href{https://orcid.org/0009-0006-1215-1473}{\includegraphics[scale=0.06]{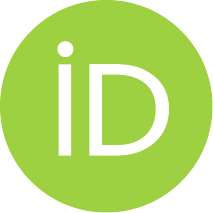}\hspace{1mm}Riku Green}\thanks{Preliminary work under review. Correspondence to riku.green@bristol.ac.uk.} \\
	School of Computer Science\\
	University of Bristol \\
    United Kingdom \\
	\texttt{riku.green@bristol.ac.uk} \\
	\And
	\href{https://orcid.org/0000-0002-8885-4443}{\includegraphics[scale=0.06]{orcid.pdf}\hspace{1mm}Grant Stevens} \\
	School of Computer Science\\
	University of Bristol \\
    United Kingdom \\
	\texttt{grant.stevens@bristol.ac.uk} \\
    \And
    {Telmo de Menezes e Silva Filho} \\
	School of Engineering Mathematics\\
	University of Bristol \\
    United Kingdom \\
	\texttt{telmo.silvafilho@bristol.ac.uk} \\
    \And
    {Zahraa Abdallah} \\
	School of Engineering Mathematics\\
	University of Bristol \\
    United Kingdom \\
	\texttt{zahraa.abdallah@bristol.ac.uk} \\
}
\else
\usepackage{authblk}

\newbox{\orcid}\sbox{\orcid}{\includegraphics[scale=0.06]{orcid.pdf}} 
\author[1]{%
	\href{https://orcid.org/0000-0000-0000-0000}{\usebox{\orcid}\hspace{1mm}David S.~Hippocampus\thanks{\texttt{hippo@cs.cranberry-lemon.edu}}}%
}
\author[1,2]{%
	\href{https://orcid.org/0000-0000-0000-0000}{\usebox{\orcid}\hspace{1mm}Elias D.~Striatum\thanks{\texttt{stariate@ee.mount-sheikh.edu}}}%
}
\affil[1]{Department of Computer Science, Cranberry-Lemon University, Pittsburgh, PA 15213}
\affil[2]{Department of Electrical Engineering, Mount-Sheikh University, Santa Narimana, Levand}
\fi

\renewcommand{\shorttitle}{\textit{DyStrat} Multistep Forecasting}

\hypersetup{
pdftitle={Time-Series Classification for  \\
           Dynamic Strategies in Multi-Step Forecasting},
pdfsubject={},
pdfauthor={Riku Green, Grant Stevens, Telmo de Menezes e Silva Filho, Zahraa Abdallah},
pdfkeywords={},
}

\begin{document}
\maketitle

\begin{abstract}
Multi-step forecasting (MSF) in time-series, the ability to make predictions multiple time steps into the future, is fundamental to almost all temporal domains. To make such forecasts, one must assume the recursive complexity of the temporal dynamics. Such assumptions are referred to as the forecasting strategy used to train a predictive model. Previous work shows that it is not clear which forecasting strategy is optimal a priori to evaluating on unseen data. Furthermore, current approaches to MSF use a single (fixed) forecasting strategy.

In this paper, we characterise the instance-level variance of optimal forecasting strategies and propose \textbf{Dy}namic \textbf{Strat}egies (DyStrat) for MSF. We experiment using 10 datasets from different scales, domains, and lengths of multi-step horizons. 
When using a random-forest-based classifier, \textbf{DyStrat outperforms the best fixed strategy}, which is not knowable a priori, \textbf{94\% of the time}, with an average \textbf{reduction in mean-squared error of 11\%}. Our approach typically triples the top-1 accuracy compared to current approaches. Notably, we show DyStrat generalises well for any MSF task.
\end{abstract}

\section{Introduction}
Multi-step forecasting (MSF) strategies have consistently received attention in the time-series literature given their necessity for long-term predictions in any dynamic domain \cite{lim2021timesurvey}. Examples where time-series forecasting is crucial range from: healthcare \cite{morid2023time_health,miotto2018deep_health}, transport networks \cite{anda2017transport, nguyen2018deep_transport}, geographical systems \cite{rajagukguk2020review_geog}, and financial markets \cite{deb2017review_energy, sezer2020financialsurvey}. Classical analysis of MSF concerns when it is appropriate to incorporate a recursive strategy or a direct strategy. The recursive strategy predicts auto-regressively on a single model's own predictions until the desired horizon length is obtained. In contrast, direct strategies require fitting many separate models which is expensive and often results in model inconsistencies \cite{taieb2014machine}. Extensive theoretical work has been done to analyse the variance-bias trade-off between these two strategies
\cite{taieb2014machine}.

\begin{figure*}[hbt!]
    \centering
    \includegraphics[width=0.95\linewidth]{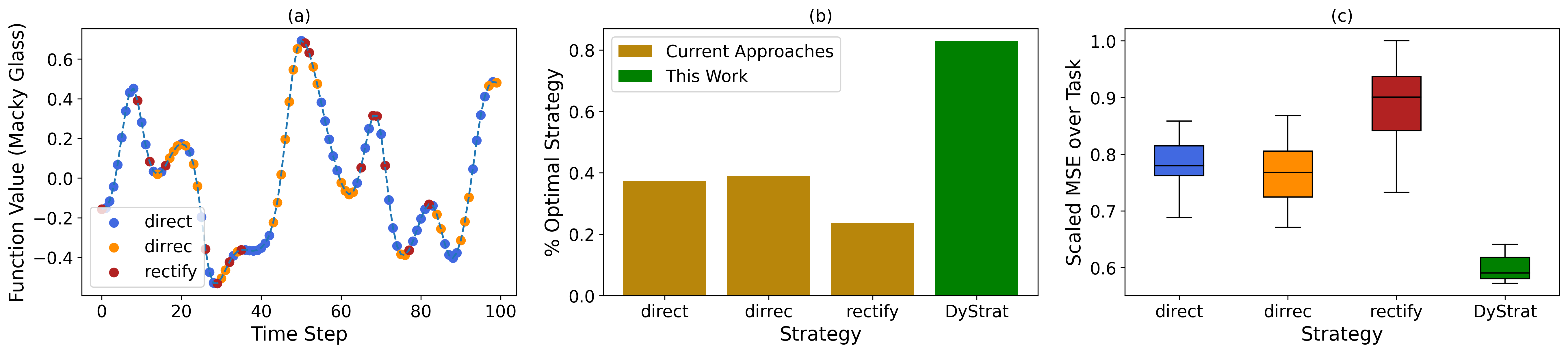}
    \caption{Current approaches to multi-step forecasting do not consider the temporal dynamics of locally optimal strategies. Therefore, they make predictions using a single forecasting strategy, such as direct, dirrec, and rectify. An example time-series (Mackey-Glass) is shown with high variance in locally optimal strategies (a), colour of dots show which example strategy performs optimally. We compare our method, DyStrat, to these current approaches using the example strategies and achieve two-fold better local optimality (b).  The error, using 1000 test points and 10 repeats, of each approach shows consistent significant error reduction using DyStrat (c).}
    \label{fig:intro}
\end{figure*}

To bridge the gap between recursive and direct strategies, \emph{hybrid strategies} have been developed \cite{taieb2012reviewmultistep}. Multi-output frameworks, such as Recursive Multi-output (RECMO) \cite{ji2017strategies} and Direct Multi-output (DIRMO) \cite{taieb2010multiple}, allow for tuning of output-dimension of models to find a `sweet spot' in bias and variance. Alternatively, DirectRecursive (dirrec) \cite{taieb2012reviewmultistep} 
and Rectify \cite{rectify} 
have been proposed.
Although hybrid methods appear to be improvements on the state-of-the-art in MSF \cite{An_comp}, the problem remains open regarding which strategy is optimal a priori to evaluating on unseen data.

We take inspiration from the success of dynamic model selection in single-step forecasting \cite{Fu_Wu_Boulet_2022}. In this work, we highlight that current approaches are strictly limiting the accuracy of forecasts by selecting forecasting strategies at the task level. To this end, we propose \textit{DyStrat}, a novel approach for MSF where strategies are dynamically selected. \autoref{fig:intro} presents an example of how current approaches are limited. There is substantial variance in strategy optimality at the instance-level (\autoref{fig:intro}a). Despite dirrec being the most locally optimal, at $\sim$40\%; DyStrat shows twice the local optimality is possible (\autoref{fig:intro}b). Lastly, optimising the local optimality translates to significant error reduction at the task level (\autoref{fig:intro}c).


We ran experiments over multiple domains of time-series data, under different lengths of forecasting horizon, sizes of training data, and forecasting model complexity. 
Our contributions are as follows:
\begin{itemize}
    \item A characterisation of the instance-level variance in strategy ranking as a MSF problem.
    \item DyStrat, a framework for MSF strategy selection as a dynamic model selection task.
    \item More accurate forecasts using DyStrat, improving previous methods by $\sim 11\%$ in forecasting error and increases top-1 strategy selection by $\sim 3$-fold.
\end{itemize}

We organise the rest of the paper as follows. Related work is discussed in \autoref{related}, followed by background (\autoref{prelims}) providing formalisation, and \autoref{method} covers the methodology. In \autoref{experiments}, we present our experiments and results, and finish with a discussion and conclusion of the paper.

\section{Related Work}
\label{related}

\textbf{Theoretical Results for Multi-step Forecasting.} The forecasting strategy represents an assumption of the recursive complexity for the data-generating process of a time-series \cite{taieb2014machine}.
In particular, it is shown that the minimisation of one-step-ahead forecast errors does not guarantee the minimum for multi-step-ahead errors, therefore the recursive strategy is asymptotically biased \cite{brown1984residual}. Whilst the direct strategy is shown to be unbiased, since the model's objective is identical to the MSF objective, there are no guarantees of consistency within the direct strategy \cite{taieb2014machine}. Another theoretically-driven strategy, \emph{rectify} \cite{rectify}, fits a recursive model to represent the majority of the dynamics and then fits a direct strategy as an asymptotically unbiased correction step. Current theoretical works are yet to consider the potential of dynamic recursive complexities.

\textbf{Empirical Results for Multi-step Forecasting.} Despite the finding that direct methods are more theoretically motivated, at least in the large data limit, it is not obvious which MSF strategy to use in practice. Multiple studies compare the performances of different MSF strategies and their findings are not entirely consistent: \citet{atiya1999comparison} favours direct strategies, \citet{taieb2012reviewmultistep} favours multi-output strategies, \citet{An_comp} favour dirrec, whereas \citet{ji2017strategies} favour recursive strategies. Further details on these strategies are covered in Section \ref{prelims}. 

Previous works do not consider dynamically selecting strategies based on instances of a task. To the best of our knowledge, they only consider using a fixed forecasting strategy on unseen data. In \citet{in2022simple}, recursive and direct forecasts weighted-averaged to create a marginally better strategy. However, the weighted strategy remains fixed for the given task. This study aims to show that a dynamic strategy leads to more accurate forecasts by removing assumptions of a fixed data-generating process.

\textbf{Multi-step Forecasting Models.} 
Naturally, work on MSF has been done using statistical forecasting models, such as ARIMA \cite{suradhaniwar2021multi, kumar2022multi}, but deep learning approaches have become state-of-the-art \cite{atiya1999comparison, taieb2012reviewmultistep, zhou2021informer}. In particular, recent works iterate transformer architectures for better performances \cite{zhou2021informer, zhou2022fedformer}, this is often attributed to transformer blocks being effective at learning long range dependencies \cite{wen2022transformers}. However, it is not always the case that transformer models outperform more simple multi-layer-perceptron approaches (MLP) \cite{xue2023make}. This can be attributed to transformers being over-parameterised and require large amounts of data.

\textbf{Time-series Classification.} 
In time-series forecasting, it has been empirically observed that classification algorithms can be used to dynamically select between a set of candidate models to produce more accurate forecasts \cite{ortega2001arbitrating}. Time-series classification (TSC) refers to the process of categorising time-series data into distinct groups \cite{bakeoff}. TSC has been successful in dynamic model selection (covered in \autoref{prelims}), where it is hypothesised that the optimal forecasting model is dependent on the local context of a time-series \cite{classifier, cruz2022two, newmetric, ortega2001arbitrating, cerqueira2017arbitrated, dmshouses_statisitical, ulrich2022classification, prudencio2004meta}. We take inspiration from these works as well as more recent work using reinforcement learning \cite{Fu_Wu_Boulet_2022} for learning contextual model selection. We identify that the literature is yet to apply these principles to the MSF task for strategy selection.



\section{Background}
\label{prelims}
We now provide formal problem definitions for multi-step forecasting and time-series classification. Given the various fields of study, we use notation consistent with the wider literature.

\textbf{Multi-step Forecasting.}
Given a univariate time-series $\mathfrak{Y} = \{\mathfrak{y}_1,...,\mathfrak{y}_T\}$ comprising $T$ observations, the goal is to forecast the next $H$ observations $\{\mathfrak{y}_{T+1},...,\mathfrak{y}_{T+H}\}$ where $H$ is the forecast horizon \cite{taieb2014machine}. In order to construct a predictive model, assumptions must be made regarding the underlying dynamics of the data-generating process for the time-series. It is common practice to use the $w$ most-recent values of the time-series as input to a predictive model. Larger $w$ allows a model to `see' further back in time. Models can be constructed by first selecting a $H$ and $w$ value, and applying a sliding-window approach over $\mathfrak{Y}$, to obtain \textit{instances}, where $x_t \in X \subseteq \Re^w$ and $y_t \in Y \subseteq \Re^H$:
\begin{equation}
    x_t = [\mathfrak{y}_{t-w},...,\mathfrak{y}_{t-1}]  \text{ and }
    y_t = [\mathfrak{y}_t,...,\mathfrak{y}_{t+H}].
    \label{slidingeq}
\end{equation}

 Following this, using the statistical learning framework \cite{vapnik_overview}, for a function family $\mathcal{H}$ parameterised over $\omega$, the function $f(x, \omega): X \rightarrow Y$ that minimises the functional: 
\begin{equation}
\label{trueRisk}
    \epsilon(\omega) = \int L(y,f(x,\omega))dP(x,y) 
\end{equation}
is considered optimal. $L(y,f(x,\omega))$ is a loss function measuring the discrepancy between the measured value $y$ and the predicted value $\hat{y}$ produced by $f$.

Since the $P(x,y)$ component in \autoref{trueRisk} is strictly unknown, optimal values for $\omega$ are found by minimising the empirical risk given by:
\begin{equation}
    \epsilon_{emp}(\omega) = \frac{1}{\ell}\sum_{i=1}^{\ell}  L(y_i, f(x_i,\omega)),
    \label{emprisk}
\end{equation}
where \autoref{emprisk} converges to \autoref{trueRisk} as $\ell$ tends to infinity \cite{vapnik_overview}. The loss function is typically chosen as the $L2$-norm. 

The mapping defined by $f:X\rightarrow Y$ can take multiple forms, and each of these are referred to as \emph{forecasting strategies}. Original work refers to the recursive and direct strategy, but it is now shown that the multi-output framework can be used to generalise these strategies \cite{ji2017strategies}.

The recursive multi-output (RECMO) strategy trains a single forecasting model, $g$. RECMO is parameterised by $\sigma$, where $\sigma$ is the number of time-steps ahead $g$ predicts for, using the sliding window from \autoref{slidingeq}. RECMO predicts recursively when $\sigma < H$, note that $\sigma$ cannot be greater than $H$. Therefore, the $n$-th recursive step of RECMO, as a function $f$, is of the form:
\begin{equation}
    f(x,\omega) = 
    \begin{cases}
    g(x), & \text{if } \sigma = H \\
    g([x_{n\sigma:w}, \hat{y}_{1:n\cdot \sigma} ]), & \text{if } \sigma < H, w>H\\
     g(\hat{y}_{H-\sigma-w:H-\sigma}), & \text{if }  \sigma < H, w<H
    \end{cases}
    \label{Recmo}
\end{equation}

The direct multi-output (DIRMO) strategy is also parameterised by $\sigma$, however it trains a set of $m$ models instead of one, $G = [g_1,...,g_m]$. Each $g_i$ in $G$ predicts $\sigma$ future values; specifically, it splits the $H$ future steps into $m$ intervals of length $\sigma$ using a similar windowing from \autoref{slidingeq} but $y_t$ begins from $\mathfrak{y}_{t+i\sigma}$ for the $g_i$. Therefore, such a forecaster $f$ when using DIRMO is of the form:
\begin{equation}
    f(x,\omega) = 
    \begin{cases}
    g(x), & \text{if } \sigma = H \\
    [g_1(x), ..., g_m(x)], & \text{if } \sigma < H
    \end{cases}
    \label{Dirmo}
\end{equation}
where $m = H/\sigma$ and $m \in \mathbb{Z}$.

Both RECMO and DIRMO are equivalent under $\sigma = H$ and become what is commonly referred to as the multi-output (MO) forecasting strategy \cite{An_comp}.

Following the formulation of multi-output strategies, direct-recursive (dirrec) \cite{An_comp} and rectify \cite{rectify} have been proposed. Dirrec trains in the same way as DIRMO, except $g_{i+1}$ has recursion of the form $g_{i+1}(x, g_i)$, where models in $G$ will use the previous models' forecasts as an input. Rectify is of the form $f_{rec} + f_{dir}$, where $f_{rec}$ denotes the recursive base model and $+f_{dir}$ denotes the correction step.

\textbf{Dynamic Model Selection.}
Dynamic model selection (DMS) is the process of learning the most suitable model among a collection of candidate-models to accurately represent a system within a specific context \cite{Fu_Wu_Boulet_2022, feng2019reinforced}. For a windowed time-series pair, $X$ and $Y$ (again from \autoref{slidingeq}), consider a set of $c$ candidate models $F = \{f_1,..., f_c\}$, where $\forall f \in F: f(X) \rightarrow Y$. Dynamic model selection aims to learn some policy, $\pi(x)$, where $\pi(x) \rightarrow \Re^{|F|}$. In the case of model ensembling, DMS uses the output of $\pi$ as the weights for the average predictions of $F$. However, if the task is restricted to binary outputs, the problem can be solved using time-series classification. In this case, $\pi$ is an indicator function such that $\pi_\theta(x) \rightarrow [0, 1]^{|F|}$; when $\pi$ is parameterised by $\theta$, the task is to minimise the following:
\begin{equation}
\label{DMS}
    \argmin_\theta (\int L(f^*(x),\pi(x,\theta))dP(x) ),
\end{equation}
where $f^*$ is the ground truth indicator of the optimal function in $F$. Optimal in this setting is the function in $F$ that minimises some desired error metric. Again, in practice, \autoref{DMS} is approximated empirically but a classification loss is used (such as cross-entropy loss). 

\section{DyStrat: Dynamic Strategy for Multi-step Forecasting}
\label{method}
In this section we outline a simple methodology for DyStrat (\textbf{Dy}namic \textbf{Strat}egies, or DS) . We propose using a time-series classification model to learn optimal multi-step forecasting strategies for given windowed time-series instances. The methodology is structured as follows: candidate strategies construction, classifier-data generation, and dynamic-strategy usage.

\textbf{Constructing Forecasters.}
The assumption behind this work follows that of DMS, where it is assumed that one MSF prediction strategy is unlikely to be optimal in all windowed instances of an MSF task. Therefore, models trained from a set of candidate forecasting strategies, $s \in S$, must be trained. We fix the function family $\mathcal{H}$ and construct $S$ by generating MSF-data using \autoref{slidingeq}. We then minimise \autoref{emprisk} over training data for all strategies and append these functions to $S$. Note that for dynamic strategies $|S| > 1$, and the DMS task difficulty increases with $|S|$ due to the output space becoming larger.

\textbf{Learning a Dynamic Strategy.} 
After $S$ is defined, we use a simple algorithm  to generate the appropriate data to learn a dynamic strategy. The same windowed pair used to construct forecasters, $X$ and $Y$, is coupled with the set $Z$, which indicates the locally optimal strategy (denoted as $s^*$) mapped from a time-series instance. We define this as $\argmin_s(L(y, s(x_i)))$;
 the procedure is shown in Algorithm \autoref{zalgo}. 

\begin{algorithm}
\caption{Procedure to Compute $Z$}
\textbf{Input:} Loss function $L$, windowed pair $X$, $Y$ of time-series

\begin{algorithmic}[1]
    \State $Z \gets []$ \Comment{Initialize an empty list for $Z$}
    \For{$i$ in $X$} \Comment{Loop over elements in $X$}
        \State $s^* \gets \text{argmin}_s(L(y, s(x_i)))$ \Comment{Compute the argmin}
        \State $Z.\text{append}(s^*)$ \Comment{Append $s^*$ to $Z$}
    \EndFor
\end{algorithmic}
\label{zalgo}
\end{algorithm}
The classifier, $\pi(x, \theta)$, is then trained to minimise: 
\begin{equation}
    \epsilon_{emp}(\theta) = \frac{1}{\ell}\sum_{i=1}^{\ell}  L(z_i, \pi(x_i,\theta)).
    \label{emppi}
\end{equation}
Importantly, the loss function when generating $Z$ in Algorithm \ref{zalgo} can be any desired metric and does not need to be differentiable. Furthermore, we acknowledge that minimising \autoref{emppi} over the same data used to train forecasting models may cause overfitting. Nonetheless, our experiments show such a method has good generalisation to unseen data.

\textbf{Using a Dynamic Strategy.} Once \autoref{emppi} is minimised, the dynamic strategy can simply be used as follows. For an unseen time-series instance, $x'$, the predicted forecast is simply given by:

\begin{algorithm}
\caption{Dynamic Strategy Procedure}
\textbf{Input:} Policy $\pi$, instance $x'$, candidate strategies $S$.
\begin{algorithmic}[1]
    \State $\hat{y}' \gets []$ \Comment{Initialize an empty list for $\hat{y}'$}

    \State $\hat{i^*} \gets \pi(x', \theta)$ \Comment{Compute predicted optimal index}
    
    \State $\hat{y}' \gets s_{\hat{i^*}}(x')$ \Comment{Compute forecast for strategy index}
    
\end{algorithmic}
\label{usage algo}
\end{algorithm}

\textbf{Best Fixed Strategy.} As mentioned before, no known strategy always performs the best. From this, we define the task-wise optimal strategy, denoted $g^* \in S$. We refer to this as \emph{the best fixed strategy}, which is not knowable a priori. Mathematically this is found by:
\begin{equation}
    \label{gstar}
     g^* =  \argmin_s \left(\frac{1}{\ell}\sum_{i=1}^{\ell}  L(y_i, s(x_i,\omega))\right).
\end{equation}
\section{Experiments}
\label{experiments}
Our experiments test the hypothesis that there is a learnable relationship between the variance in optimal strategy of an MSF task and the windowed instances. 

\textbf{Datasets.} We conduct analysis on multiple well-studied datasets: the Mackey-Glass (MG) equations, which represent a chaotic system that models biological processes \cite{glass2010mackey}, the Electricity Transformer Temperature (ETT) dataset (m1, m2, h1, and h2 versions) from \citet{zhou2021informer}, the PEMS8 traffic dataset from \citet{guo2019attention}, sunspots dataset from \citet{pala2019forecastingsun}, Poland's energy consumption from \citet{sorjamaa2007methodology}, and two more synthetic datasets (Lorenz system \cite{frank2001timelorenz} and 5\% noise sine wave). In total, three synthetic datasets and seven real-world datasets are used in this study. Dataset lengths are in \autoref{Datatable} and their values are normalised between zero and one.

\begin{table}[hbt!]
\centering
\begin{tabular}{@{}lll@{}}
\toprule
Dataset      & Length & Type       \\ \midrule
Mackey Glass \cite{glass2010mackey} & 10,000 & Synth  \\
ETT m1  \cite{zhou2021informer}     & 69,680 & Real \\
ETT m2  \cite{zhou2021informer}     & 69,680 & Real \\
ETT h1  \cite{zhou2021informer}     & 17,420 & Real \\
ETT h2  \cite{zhou2021informer}     & 17,420 & Real \\
PEMS08  \cite{guo2019attention}     & 17,856 & Real \\
Sunspots \cite{pala2019forecastingsun}    & 3,265  & Real \\
Poland  \cite{sorjamaa2007methodology}     & 1,465  & Real \\
Lorenz  \cite{frank2001timelorenz}     & 10,000 & Synth  \\
Sine (5\% gaussian noise)         & 10,000 & Synth  \\ \bottomrule
\end{tabular}
\caption{Datasets used in this study, length, and real-world (real) / synthetic (synth).}
\label{Datatable}
\end{table}


\textbf{Candidate Strategies.} We provide a brute-force analysis to rigorously test our claim that dynamic strategies are superior to fixed ones. For a multi-step-ahead horizon of $H$, the number of possible $\sigma$ parameters is the number of divisible factors in $H$. We consider all possible RECMO and DIRMO strategies in our experiments. 
We also include Rectify and dirrec, which are not parameterised by $\sigma$.

\textbf{Forecasting Functions.} We use a multi-layer perceptron (MLP), with a single hidden layer, as the hypothesis space for forecasting functions. This is for three reasons: they perform well on MSF \cite{xue2023make}, they are easy to train, and their complexity can be easily adjusted by varying the hidden layer width. We use Python's SKlearn \cite{scikit-learn} package with its in-built MLPRegressor class with default hyperparameters; we only vary the 
hidden-layer hyper-parameter (larger hidden layer corresponds to a more complex function class).

\textbf{Time-series classifiers.}
We compare four different classification function-families for $\pi$: a linear classifier, MLP, K-nearest neighbours (KNN), and a time-series random-forest (TSF). We include the linear classifier as a high bias learner and the MLP as a high variance counterpart. The KNN is used as in \citet{yu2022dynamic}, and TSF as a time-series-specific classifier from the python library PyTS \cite{tsforst}. We expect TSF to perform best as it is the only classifier specifically designed for learning temporal patterns.

\textbf{Task Settings.}
We vary $H \in [10, 20, 40, 80, 160]$ to test dynamic strategies across long and short term horizons. Similarly, we test the effect in changing scale of training-data by varying $\ell_\% \in [10, 20, 40, 80]$, where $\ell_\%$ denotes the percentage of data used to train forecasters.
Default values, unless stated otherwise, are: $n\_hidden\_layer = 100$, $\ell_\% = 75$, $W = 2$, and $H = 20$. We use 10\% of all available data for each dataset as unseen data for evaluation. 
We also include a sensitivity study on MG data where we vary the hidden layer width and feature window length. Tasks are repeated 5 times with mean and standard deviations shown.

\textbf{Evaluation Metrics.} We consider two main metrics to compare DyStrat with previous approaches.
Mean-squared error is used since this is equivalent to the loss functions training forecasters to maintain consistency. Often we reference  $g^*$ (\autoref{gstar}) as a comparison to the best current approach. We also show the ranking error of strategies; this is a scale-free metric and highlights a statistic on how each strategy performs in direct comparison with each other. We present results from other metrics in the literature (MAE, MAPE, and SMAPE) in \autoref{othermetrics}, as well as raw errors in \autoref{raw mse results}. From 
\autoref{emppi}, we evaluate the error of the optimal \emph{dynamic strategy} at test time and divide all errors by this optimal error. This lower bounds other strategy-errors by $1$ and makes comparison across both DyStrat and previous approaches easier. 
Additionally, from Algorithm \ref{zalgo} to construct $Z$, the set of $s^*$, we can easily compute the top-1 accuracy of a strategy as the proportion of $Z$ that indexes that strategy.
Ranking metrics give credit to strategies that do not rank optimally at the instance-level, whereas top-1 only gives credit if a strategy is optimal.
\begin{figure}[t!]
    \centering
    \includegraphics[width = \linewidth]{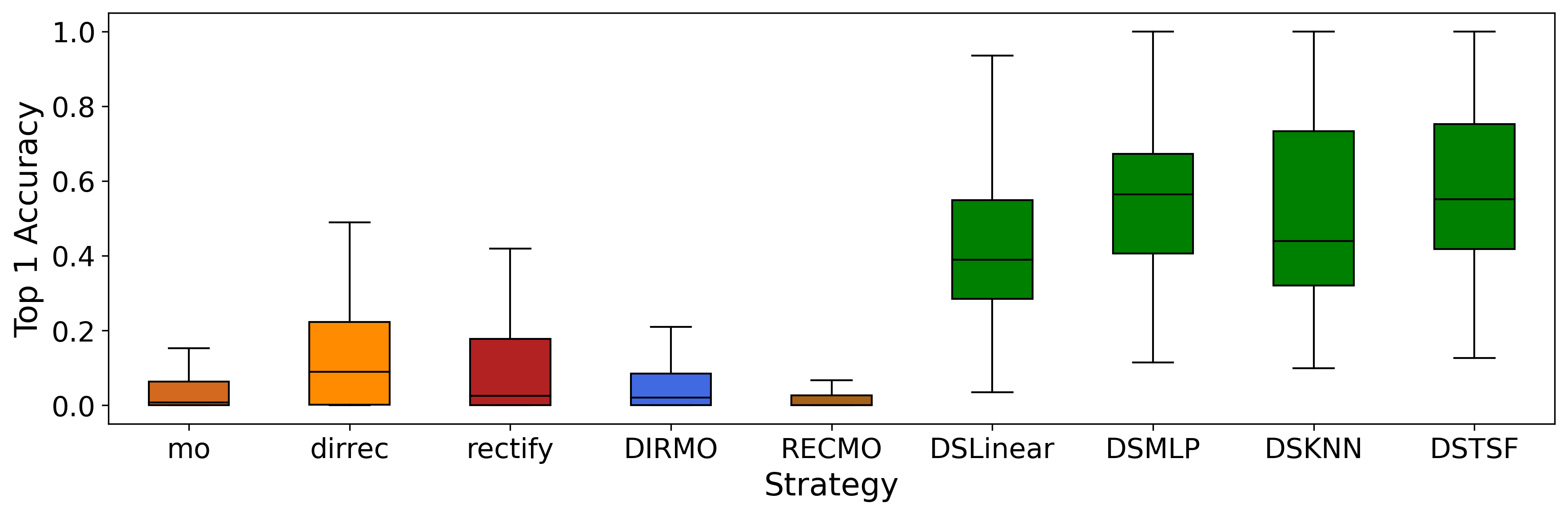}
    \caption{The top-1 accuracy (the proportion of within-task instances where a strategy is optimal) aggregated over all datasets and task settings. DIRMO and RECMO include all $\sigma$ parameters.}
    \label{fig:top1}
\end{figure}

\textbf{Task-level and Instance-level comparison.} A new perspective for comparing strategies on the MSF task is the instance-level analysis. When discussing `task-level' ranks, this is done by evaluating the ranks over aggregated statistics of the entire test dataset. On the other hand, `instance-level' ranks refer to evaluating the ranks of strategies per instance of the test dataset and then aggregating these values. Task-level ranks demonstrate how strategies rank on the overall error, and instance-level ranks demonstrate how strategies are expected to rank on a given instance. As an example, a strategy may achieve the lowest instance-level rank, but not achieve the lowest task-level rank. This can be attributed to ranking being a scale-free metric. Nonetheless, ranking metrics reveal how strategies compete at each level. Since DyStrat selects from $S$, we use the \emph{dense ranking} to ensure that the instance-level ranks are bounded between $1$ and $|S|$.

\begin{table*}[hbt!]
\centering
\fontsize{7}{12}
\begin{tabular}{@{}>{\footnotesize}ccccccccccccccc@{}}
\toprule
Dataset & mo                & rc               & d1               & r1               & dirrec              & d2               & r2               & d4               & r4                & d5               & r5                & d10               & \textbf{DyStrat}               \\ \midrule
ETTh1   & \textit{1.9±0.1}  & \textit{3.2±0.6} & {\ul \textit{1.9±0.0}} & \textit{2.3±0.6} & \textit{2.4±0.2} & \textit{2.1±0.2} & \textit{2.2±0.4} & \textit{1.9±0.1} & \textit{2.1±0.2}  & \textit{2.1±0.2} & \textit{2.0±0.1}  & \textit{1.9±0.0}  & {{\textbf{1.7±0.1}}} \\
ETTh2   & \textit{4.5±0.4}  & {\ul \textit{1.4±0.2}} & \textit{2.2±0.1} & \textit{2.0±0.4} & \textit{2.1±0.2} & \textit{2.6±0.2} & \textit{3.9±1.8} & \textit{3.4±0.6} & \textit{4.8±0.3}  & \textit{3.4±0.3} & \textit{4.8±0.7}  & \textit{5.7±0.4}  & {{\textbf{1.2±0.0}}} \\
ETTm1   & \textit{2.0±0.4}  & \textit{4.1±1.8} & \textit{2.6±0.6} & \textit{2.5±0.7} & \textit{3.1±0.6} & \textit{2.6±0.4} & \textit{2.6±0.4} & \textit{2.4±0.5} & \textit{2.8±0.7}  & \textit{2.2±0.4} & {\ul \textit{1.7±0.4}}  & \textit{1.9±0.4}  & {\textbf{1.4±0.2}} \\
ETTm2   & \textit{$\geq$10} & \textit{2.9±0.6} & \textit{2.7±0.8} & \textit{4.2±4.1} & {\ul \textit{2.3±0.7}} & \textit{3.2±0.2} & \textit{5.9±2.4} & \textit{6.2±1.9} & \textit{$\geq$10} & \textit{6.8±0.9} & \textit{8.6±1.1}  & \textit{$\geq$10} & {{\textbf{1.2±0.1}}} \\
PEMS8   & \textit{2.3±0.2}  & \textit{2.0±0.1} & {\ul \textit{1.8±0.3}} & \textit{5.4±1.8} & \textit{2.2±0.5} & \textit{1.8±0.2} & \textit{8.7±2.1} & \textit{2.2±0.5} & \textit{$\geq$10} & \textit{2.0±0.2} & \textit{$\geq$10} & \textit{1.9±0.0}  & {{\textbf{1.3±0.1}}} \\ \bottomrule
\end{tabular}

\caption{DyStrat using TSForest is compared to fixed strategies on unseen data for 5 benchmark datasets with \textbf{10\% training data for forecasters}
. Headings show the forecasting strategy used and datasets are shown on the left. Values show the MSE relative to the optimal strategy,`$\geq 10$' is used for large errors. {\ul Underline} shows best baseline and \textbf{bold} shows best method. Results over 5 random seeds.}
\label{alloverL10}
\end{table*}

\begin{table*}[hbt!]
\centering
\fontsize{7}{12}
\begin{tabular}{@{}>{\footnotesize}ccccccccccccccc@{}}

\toprule
Dataset & mo               & rc               & d1               & r1                 & dirrec               & d2                 & r2                        & d4               & r4                & d5               & r5               & d10              & \textbf{ DyStrat}               \\ \midrule
ETTh1   & \textit{2.1±0.4} & \textit{9.8±1.2} & \textit{5.1±0.6} & \textit{4.8±0.9}   & \textit{6.1±0.2}  & \textit{4.0±0.2}   & \textit{3.4±1.3}          & \textit{3.4±0.1} & \textit{3.1±0.2}  & \textit{2.8±0.1} & \textit{4.1±1.2} & {\ul \textit{1.3±0.2}} & {\textbf{1.2±0.1}} \\
ETTh2   & \textit{2.2±0.3} & \textit{4.0±1.1} & \textit{2.7±0.2} & \textit{3.8±1.7}   & \textit{2.8±0.3}  & \textit{2.8±0.6}   & \textit{3.8±0.2}          & \textit{3.0±0.7} & \textit{4.5±1.8}  & \textit{2.4±0.7} & \textit{4.7±1.3} & {\ul \textit{2.1±0.1}} & {\textbf{1.4±0.1}} \\
ETTm1   & \textit{$\geq$10}       & \textit{$\geq$10}       & \textit{$\geq$10}       & \textit{6.8±0.6}   & \textit{$\geq$10}        & \textit{$\geq$10} & {\ul \textit{2.4±0.5}} & \textit{$\geq$10}       & \textit{$\geq$10}        & \textit{$\geq$10}       & \textit{$\geq$10}       & \textit{$\geq$10}       & {\textbf{1.2±0.0}} \\
ETTm2   & {\ul \textit{1.8±0.3}} & \textit{6.1±0.6} & \textit{8.8±2.2} & \textit{$\geq$10} & \textit{11.1±1.0} & \textit{5.1±0.3}   & \textit{5.3±1.2}          & \textit{2.2±0.1} & \textit{6.2±6.4}  & \textit{2.3±0.6} & \textit{2.1±0.9} & \textit{2.3±0.5} & {\textbf{1.3±0.1}} \\
PEMS8   & \textit{2.1±0.1} & \textit{1.6±0.2} & {\ul \textit{1.6±0.1}} & \textit{9.2±1.0}   & \textit{1.6±0.0}  & \textit{1.6±0.1}   & \textit{$\geq$10}         & \textit{1.8±0.2} & \textit{$\geq$10} & \textit{1.8±0.1} & \textit{7.6±1.2} & \textit{2.1±0.1} & {\textbf{1.3±0.0}} \\ \bottomrule
\end{tabular}

\caption{DyStrat using TSForest is compared to fixed strategies on unseen data for 5 benchmark datasets with \textbf{80\% training data for forecasters}. Headings show the forecasting strategy used, and datasets are shown on the left. Values show the MSE relative to the optimal strategy,`$\geq 10$' is used for large errors. {\ul Underline} shows best baseline and \textbf{bold} shows best method. Results over 5 random seeds.}
\label{alloverL80}
\end{table*}
\begin{figure*}[hbt!]
    \centering
    \begin{minipage}{.49\linewidth}
        \centering
        \includegraphics[width=\linewidth]{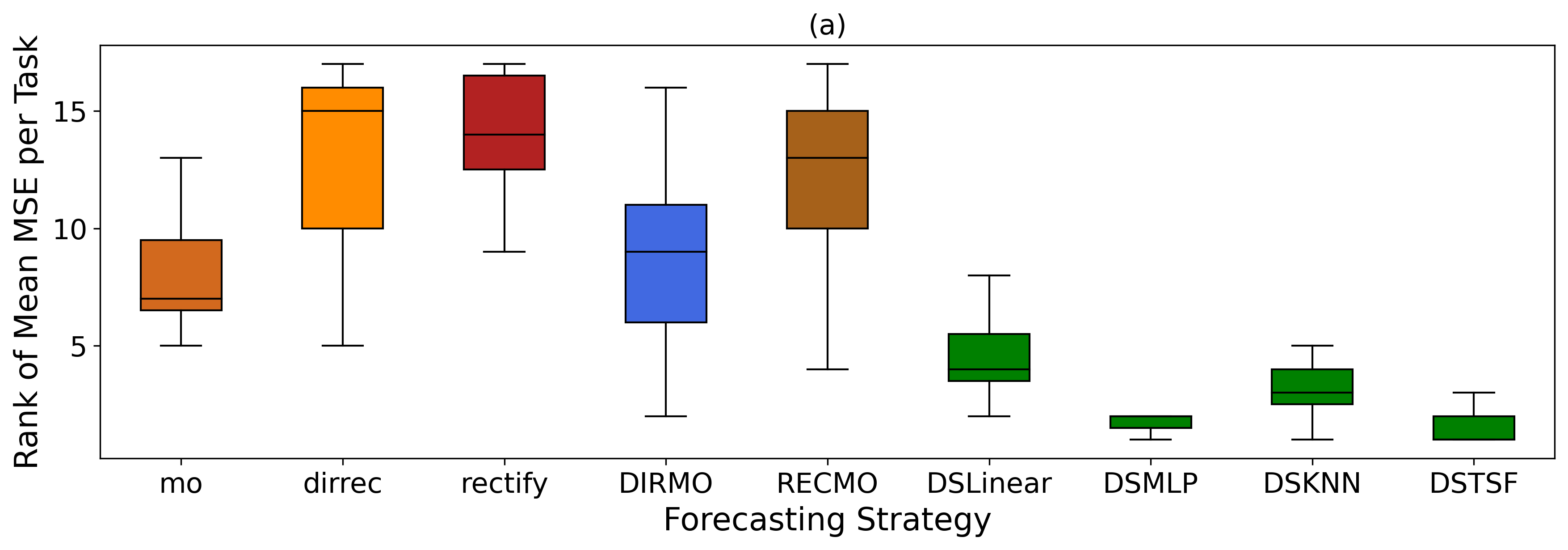}
    \end{minipage}\hfill
    \begin{minipage}{.49\linewidth}
        \centering
        \includegraphics[width=\linewidth]{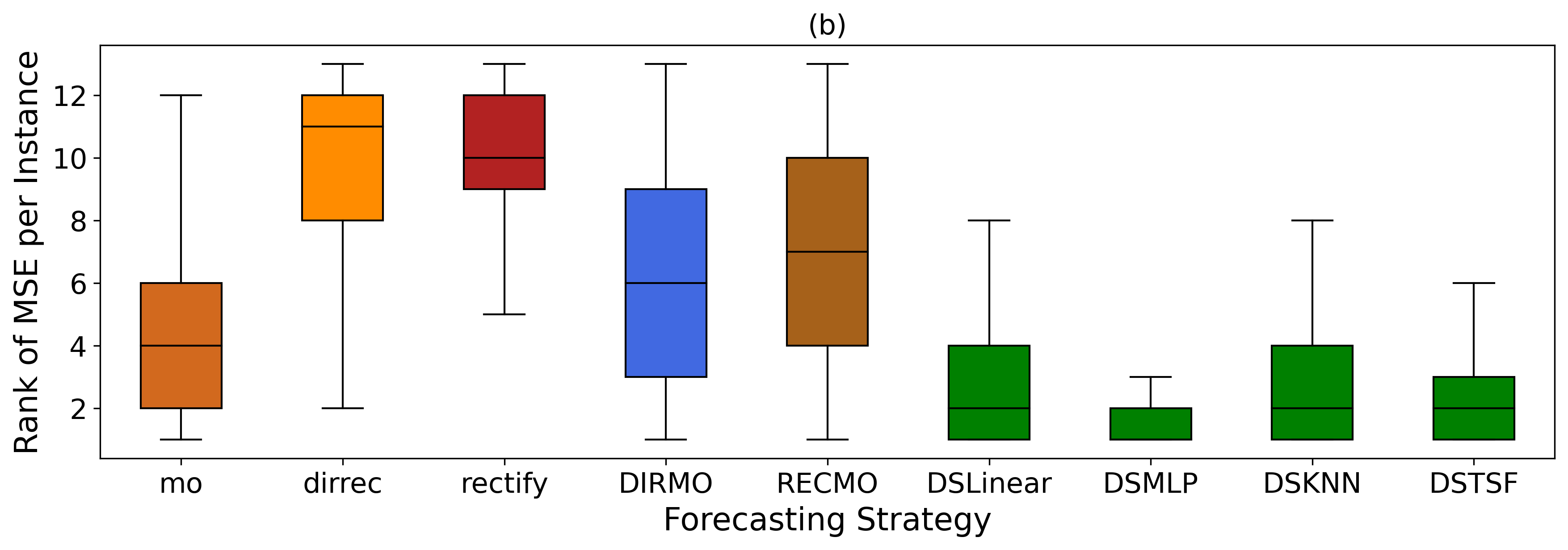}
    \end{minipage}
  
    \caption{Raw MSE from \autoref{alloverL80} is mean-averaged and ranked per-task for each dataset (a). Raw MSE from \autoref{alloverL80} is ranked per-instance and mean-averaged for each dataset (b). DS$C$ is DyStrat using $C$ as the classifier.}
    \label{fig:withinoverL}
\end{figure*}


\subsection{Results}
\label{results}


We present results using the relative error of strategies to the optimal \textit{dynamic strategy} of each task for $\ell_\% = 10$ and $80$ (\autoref{alloverL10} and \ref{alloverL80}), and $H = 10$ and $160$ (\autoref{alloverH10} and \ref{alloverH160}). The remaining settings in \autoref{raw mse results}, and a summary of remaining metrics in \autoref{othermetrics}. 

\textbf{DyStrat improves performance independent of training-data size.} We show the performance of DyStrat on all ETT datasets and PEMS08 data against all benchmarks in \autoref{alloverL10} and \autoref{alloverL80} (horizon length 20). DyStrat outperforms every benchmark in both 10\% and 80\% training data regimes with 9\% and 18\% reduction in error, respectively compared to $g^*$. \autoref{raw mse results} shows the remaining results for performance at intermediate percentages of training data. We find that DyStrat is effective in all sizes of training data considered. 


We also show the ranking performance of strategies from \autoref{alloverL80} in \autoref{fig:withinoverL}. This is done at the task level, where the overall MSE on a task is ranked (\autoref{fig:withinoverL}a), and also at the instance level, where the MSE per instance is ranked and then the mean-average is taken (\autoref{fig:withinoverL}b). All DyStrat approaches outperform the benchmarks in ranking over tasks. We explain this result by the instance-level ranking performance, where it is clear that the DyStrat approaches learn to correctly predict top-ranking strategies. The accumulation of these predictions results in lower error on the task level.

\begin{table*}[hbt!]
\centering
\fontsize{10}{12}
\begin{tabular}{@{}>{\footnotesize}ccccccccccccccc@{}}

\toprule
Dataset & mo                        & rectify                   & d1                        & r1               & dirrec                       & d2                        & r2               & d5                        & r5                        & \textbf{DyStrat}     \\ \midrule
ETTh1   & {\ul \textit{1.9±0.4}}          & \textit{4.9±1.7}          & \textit{3.7±1.0}          & \textit{2.9±0.4} & \textit{4.5±0.3}          & \textit{3.9±1.3}          & \textit{4.2±0.9} & \textit{2.8±0.4}          & \textit{3.2±0.9}          & {\textbf{1.6±0.2}} \\
ETTh2   & {\ul \textit{2.6±0.2}}          & \textit{5.0±0.6}          & \textit{2.9±0.7}          & \textit{5.9±1.5} & \textit{3.5±0.7}          & \textit{3.6±0.8}          & \textit{7.0±2.1} & \textit{3.5±0.5}          & \textit{4.0±1.3}          & {\textbf{1.6±0.1}} \\
ETTm1   & \textit{\textgreater{}10} & \textit{\textgreater{}10} & \textit{\textgreater{}10} & {\ul \textit{4.7±2.6}} & \textit{\textgreater{}10} & \textit{\textgreater{}10} & \textit{6.6±6.6} & \textit{\textgreater{}10} & \textit{7.1±7.6}          & {\textbf{1.2±0.0}} \\
ETTm2   & \textit{3.6±1.5}          & \textit{5.2±0.8}          & \textit{9.2±3.8}          & \textit{5.6±2.9} & \textit{\textgreater{}10} & \textit{6.2±4.3}          & \textit{2.8±1.2} & {\ul \textit{2.0±0.8}}          & \textit{2.6±1.1}          & {\textbf{1.3±0.1}} \\
PEMS8   & \textit{3.0±0.3}          & {\ul \textit{1.5±0.1}}          & \textit{2.0±0.1}          & \textit{9.9±2.0} & \textit{2.1±0.2}          & \textit{2.1±0.2}          & \textit{7.6±1.7} & \textit{2.7±0.2}          & \textit{\textgreater{}10} & {\textbf{1.4±0.0}} \\ \bottomrule
\end{tabular}
\caption{DyStrat using TSForest is compared to fixed strategies on unseen data for 5 benchmark datasets with \textbf{multi-step horizon of 10}. Headings show the forecasting strategy used, and datasets are shown on the left. Values show the MSE relative to the optimal strategy,`$\geq 10$' is used for large errors. {\ul Underline} shows best baseline and \textbf{bold} shows best method. Results over 5 random seeds.}
\label{alloverH10}
\end{table*}

\begin{table*}[hbt!]
\centering
\fontsize{9}{12}
\begin{tabular}{@{}>{\footnotesize}ccccccccccccccc@{}}
\toprule
Dataset & mo                        & rectify                   & d1                        & r1                        & d32                       & r32              & d40              & r40              & d80              & r80              & \textbf{DyStrat}     \\ \midrule
ETTh1   & \textit{\textgreater{}10} & \textit{\textgreater{}10} & \textit{\textgreater{}10} & \textit{\textgreater{}10} & \textit{7.7±0.6}          & \textit{8.6±1.6} & \textit{5.2±0.6} & \textit{6.4±0.7} & {\ul \textit{1.4±0.0}} & \textit{1.9±0.5} & {\textbf{1.4±0.0}} \\
ETTh2   & \textit{2.5±0.2}          & \textit{8.6±0.4}          & \textit{3.9±0.2}          & \textit{6.7±2.2}          & {\ul \textit{1.3±0.1}} & \textit{1.9±0.1} & \textit{1.4±0.2} & \textit{1.4±0.1} & \textit{3.6±0.4} & \textit{3.6±0.1} & {\textbf{1.2±0.1}} \\
ETTm1   & \textit{3.8±0.4}          & \textit{\textgreater{}10} & \textit{\textgreater{}10} & \textit{\textgreater{}10} & \textit{2.3±0.2}          & \textit{2.0±0.2} & \textit{2.2±0.3} & {\ul \textit{1.7±0.1}} & \textit{2.0±0.2} & \textit{2.4±0.5} & {\textbf{1.3±0.0}} \\
ETTm2   & \textit{2.2±0.2}          & \textit{4.7±0.2}          & \textit{6.0±0.4}          & \textit{8.9±3.3}          & {\ul \textit{1.5±0.0}}          & {\ul \textit{1.5±0.0}} & \textit{2.4±0.6} & \textit{2.1±0.2} & \textit{7.2±0.3} & \textit{7.6±1.4} & {\textbf{1.0±0.0}} \\
PEMS8   & \textit{1.5±0.1}          & \textit{1.7±0.2}          & \textit{1.2±0.0}          & \textit{\textgreater{}10} & \textit{1.5±0.0}          & \textit{7.1±1.1} & \textit{1.4±0.0} & \textit{7.0±0.9} & {\ul \textit{1.3±0.1}} & \textit{5.2±0.9} & {\textbf{1.1±0.0}} \\ \bottomrule
\end{tabular}
\caption{DyStrat using TSForest is compared to top 6 fixed strategies on unseen data for 5 benchmark datasets with \textbf{multi-step horizon of 160}. Headings show the forecasting strategy used and datasets are shown on the left. Values show the MSE relative to the optimal strategy,`$\geq 10$' is used for large errors.  {\ul Underline} shows best baseline and \textbf{bold} shows best method. Results over 5 random seeds.}
\label{alloverH160}
\end{table*}


\begin{figure*}[hbt!]
    \centering
    \begin{minipage}{0.48\textwidth}
        \centering
        \includegraphics[width=\linewidth]{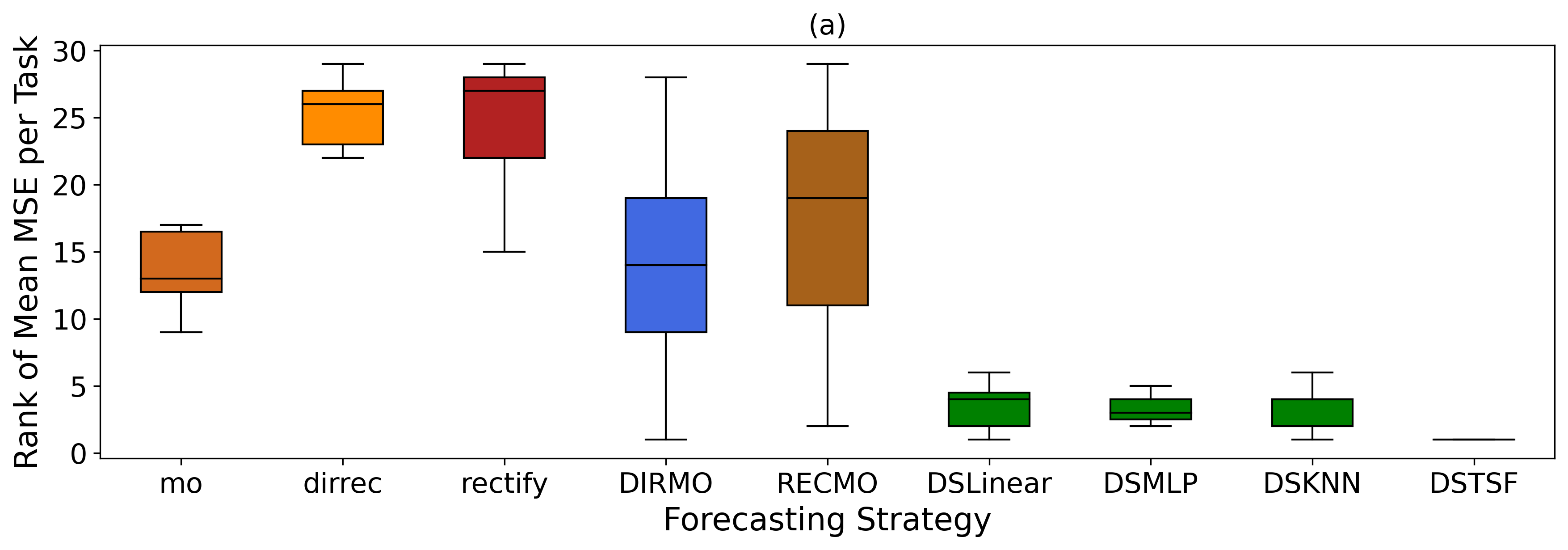}
    \end{minipage}\hfill
    \begin{minipage}{0.48\textwidth}
        \centering
        \includegraphics[width=\linewidth]{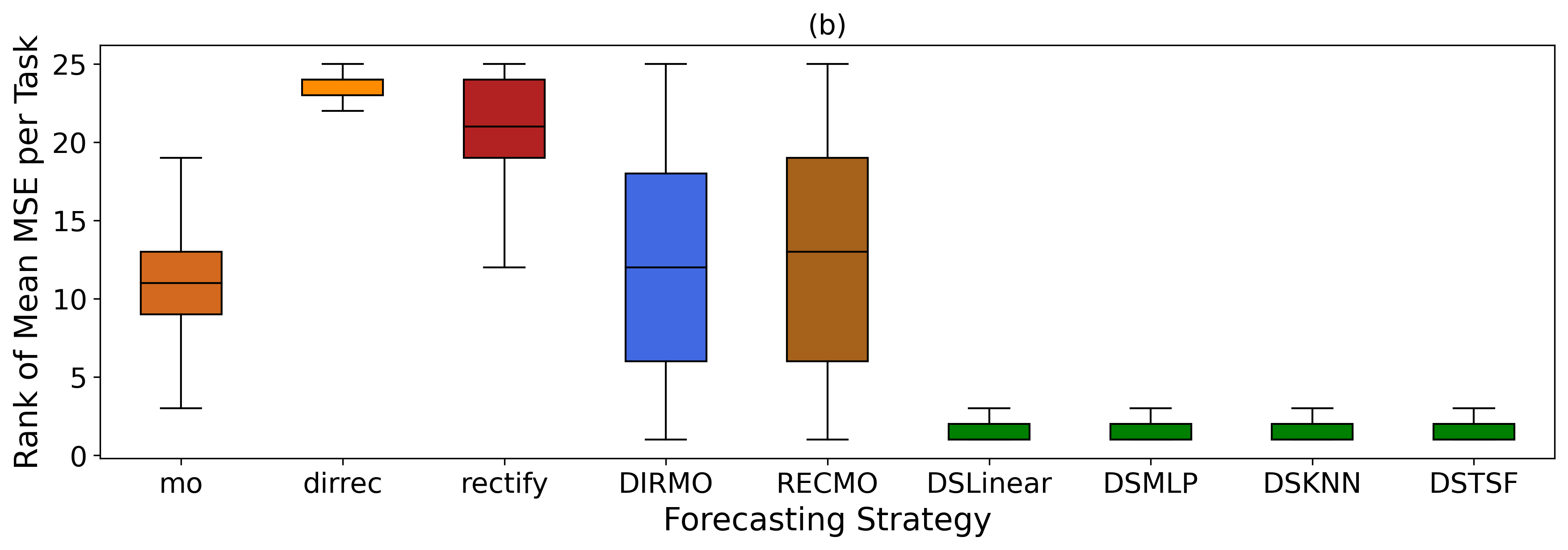}
    \end{minipage}
    \caption{Raw MSE from \autoref{alloverH160} is mean-averaged and ranked per-task for each dataset (a). Raw MSE from \autoref{alloverH160} is ranked per-instance and mean-averaged for each dataset (b). DS$C$ is DyStrat using $C$ as the classifier.}
    \label{fig:withinoverH}
\end{figure*}

\textbf{DyStrat improves performance independent of multi-step horizon length.} We fix the size of training data to 75\%, with \autoref{alloverH10} showing performance on a short-horizon of 10 and \autoref{alloverH160} on a long-horizon of 160; DyStrat using TSF reduces error by 24\% and 9\%, respectively compared to $g^*$. 


Again, we show the task-level and instance-level ranking performance analysis as before, for horizons 10 and 160 (\autoref{fig:withinoverH} a and b, respectively). Given the brute-force approach of this study, increasing the horizon length increases the number of candidate strategies, which makes the classification task harder. Despite this, DyStrat learns an effective instance-level ranking (\autoref{fig:withinoverH}b) and consistently ranks better than all other strategies (\autoref{fig:withinoverH}a).

\textbf{Better Top-1 Accuracy with DyStrat.} We record the proportion of instances where strategies are rank-one (top-1 accuracy) per task of this study in \autoref{fig:top1}. We find that dirrec is generally most often acting as $s^*$ for a given task ($\sim$ 12\%). DyStrat approaches achieve between 40-58\% top-1 accuracy, in contrast. These results strongly support the hypothesis that the relationship between instances and $s^*$ generalises well to unseen data. DyStrat using a simple linear classifier is twice as accurate as dirrec, showing that often times this relationship is simple to learn.

\begin{figure}[hbt!]
    \centering
    \includegraphics[width=\linewidth]{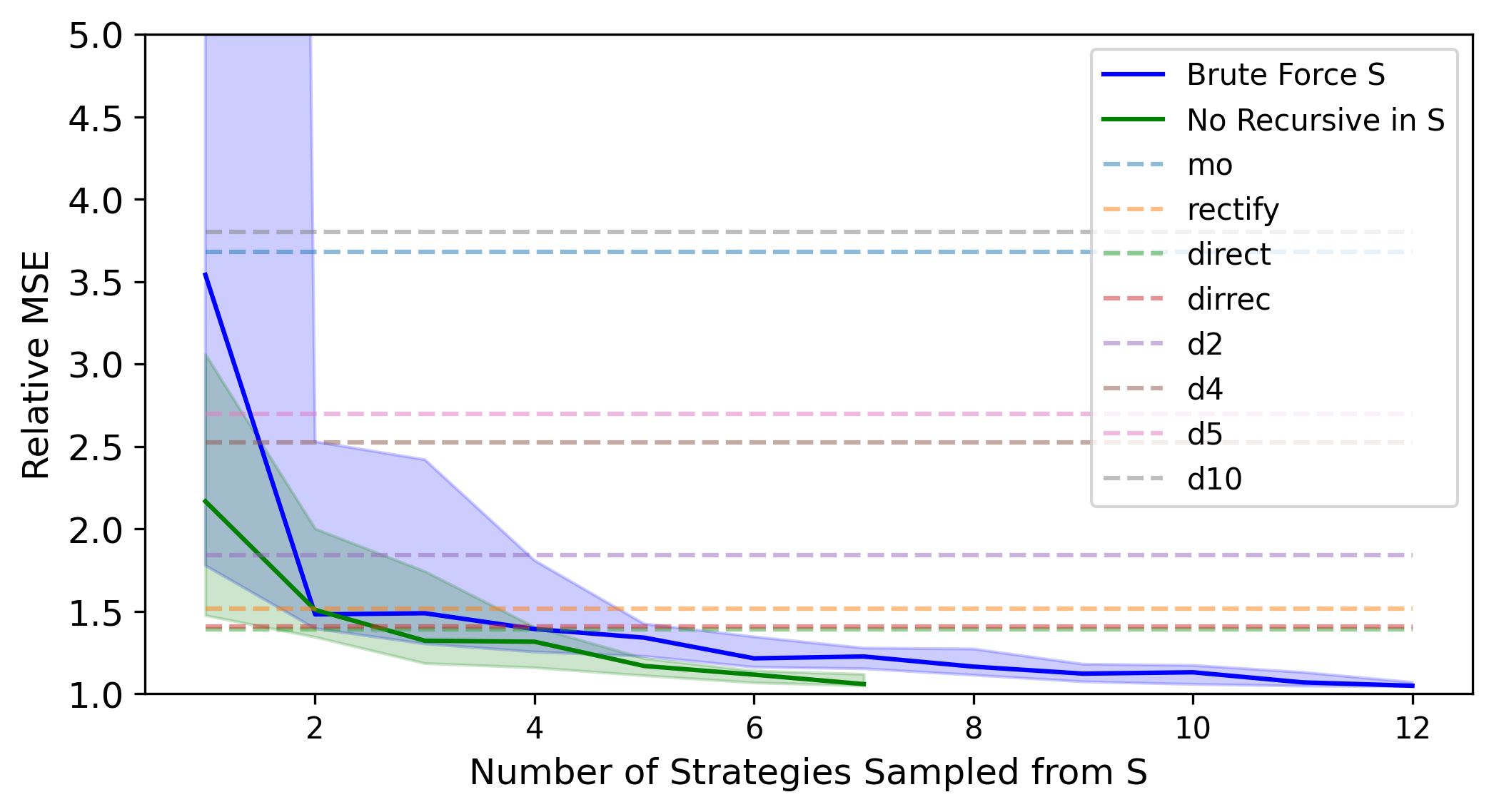}
    \caption{DyStrat is given all possible candidate strategies (brute force in blue) and a restricted set that excludes recursive strategies (green). We sample strategies uniformly and plot their relative error to the optimal dynamic strategy. Lines show medians and shaded regions show interquartile range across 30 runs.}
    \label{fig:sampled}
\end{figure}

\textbf{Quan{\ul t}ity Over Qua{\ul l}ity.} In practice, DyStrat requires deciding which known strategies to include in $S$. We compare the performance of DyStrat when $S$ only includes direct strategies (which are best performing on this task), against the full brute-force strategies $S$. This comparison is shown in \autoref{fig:sampled} and has two findings. Firstly, DyStrat is not negatively affected by strategies that rank poorly at the task level; in fact, it is able to utilise them to achieve a lower task-level error. This is shown by the blue line achieving a lower error than the green line ($1.06$ vs $1.19$) in \autoref{fig:sampled}. We find that errors generally decrease as $|S|$ increases, with a sharp reduction in error even when selecting only between two strategies. This is significant because TSF practitioners can expect reductions in task error without requiring large $|S|$ nor selective filtering of its elements.

\textbf{Forecasting-function complexity invariance.} We vary the hidden layer width of the MLP forecaster used. This analysis is only done on MG data as a case study. \autoref{fig:widthsense} (top) shows that the dynamic strategies using TSF and KNN are consistently much better than $g^*$ with near-optimal relative error, and with low variance in performance. Therefore, regardless of the computational constraints of practitioners for training forecasting models, DyStrat is offers improved performance for MSF tasks.

\textbf{Feature-window length invariance.} We vary the length of the feature window for training forecasting models, shown in \autoref{fig:widthsense} (bottom). DyStrat significantly outperforms $g^*$, with near-optimal relative error and has much lower variance in error; again, this is shown only for MG data. Given that dataset size often constrains the feature window length, DyStrat provides a performance boost to data of varying scales and so is applicable to any domain.

\textbf{Ablation of best fixed strategy independent.} \autoref{Ablation table} compares the performance of the optimal fixed strategy to a dynamic one with access to all candidate strategies \emph{except} for the optimal one. This finding suggests that even misspecifying the candidate strategies, such that $g^*$ is removed, often results in more accurate forecasts when using a dynamic strategy. It appears that advancing dynamic-strategy methods holds more promise for research progress than the pursuit of increasingly complex strategies.

\begin{figure}[hbt!]

  \begin{subfigure}
    \centering
    \includegraphics[width=0.95\linewidth]{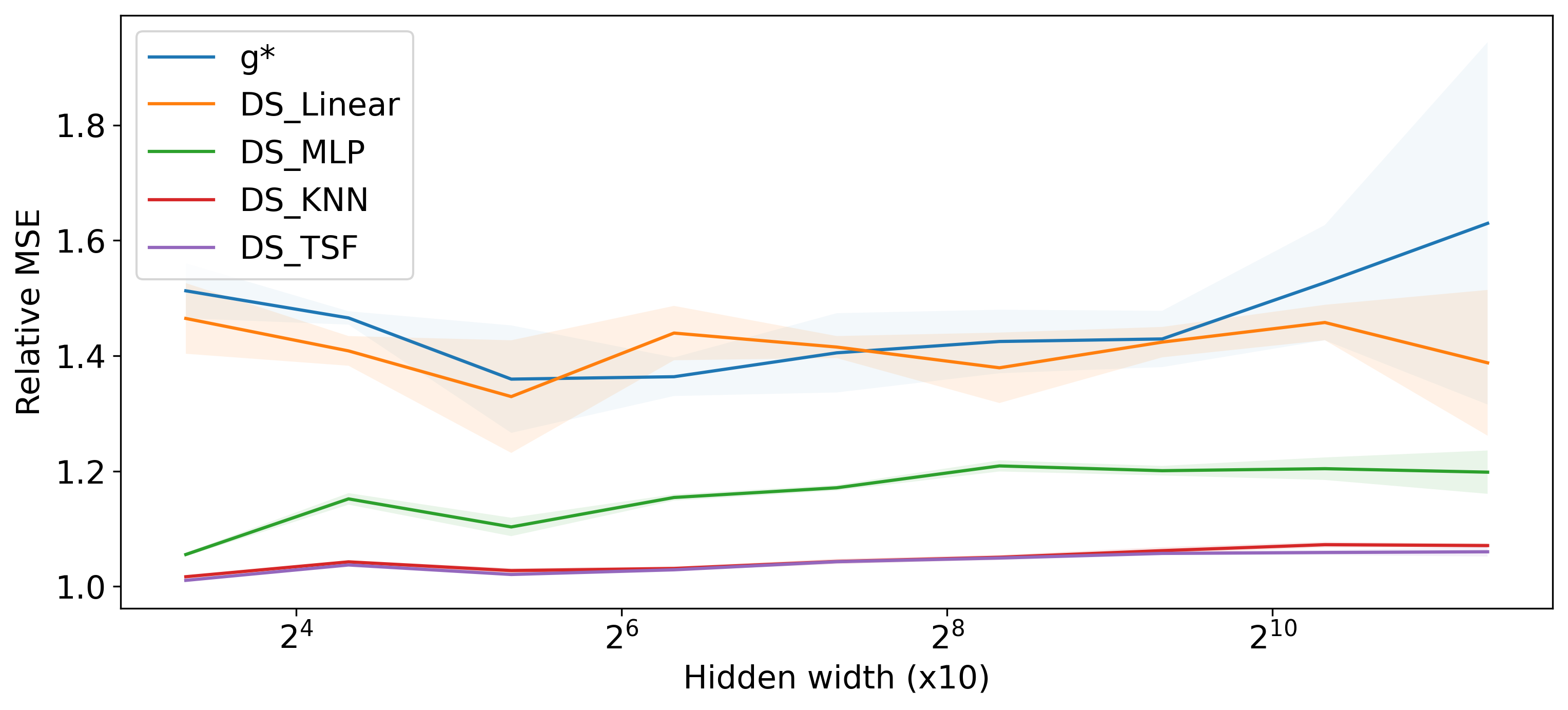}
    \label{fig:widthsense}
    \end{subfigure}
    
  \begin{subfigure}
    \centering
    \includegraphics[width=0.95\linewidth]{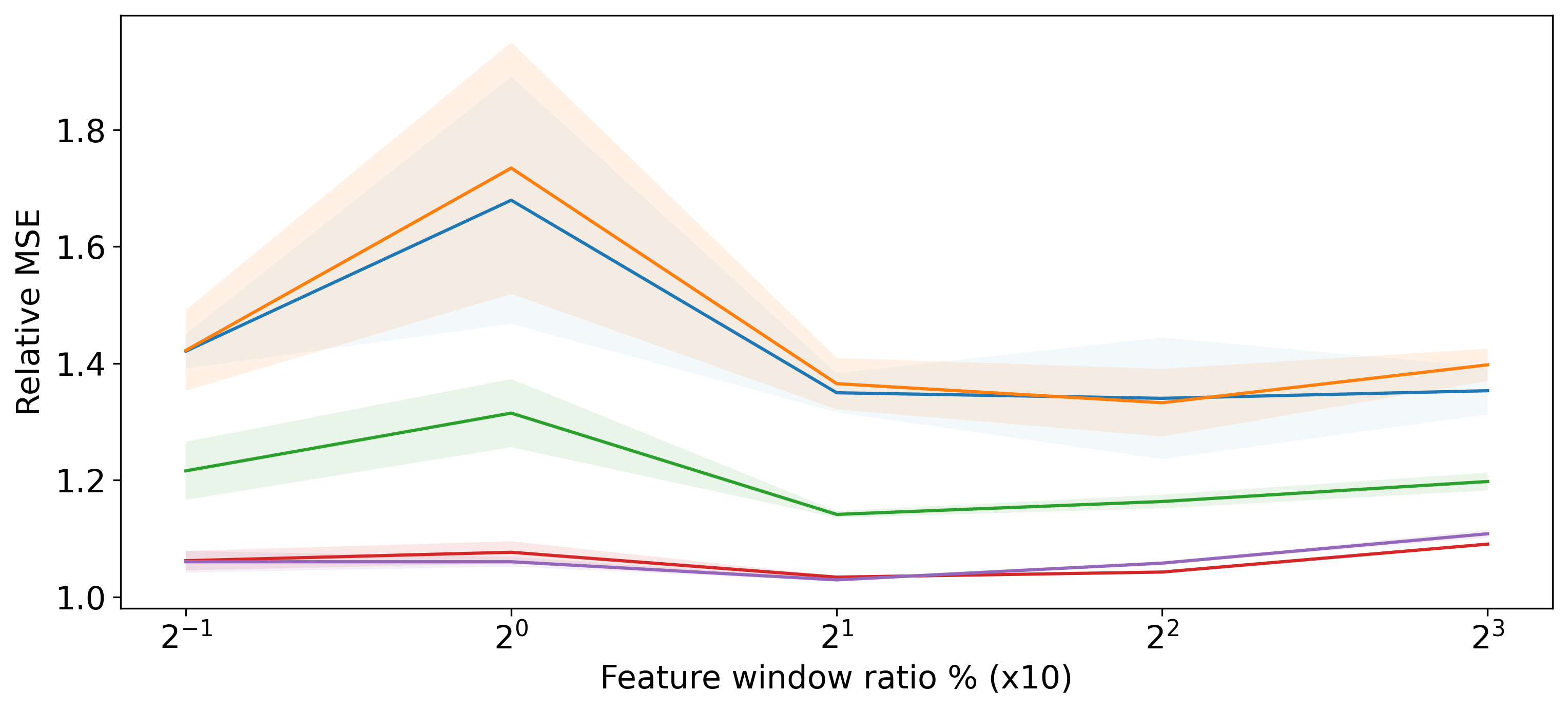}
    \label{fig:windowsense}
  \end{subfigure}

  \caption{The relative MSE of the best fixed strategy over the task ($g^*$) is compared to those of learnt dynamic strategies. We vary the complexity of forecasters by varying hidden layer width (a), and the width of the feature window of the forecaster (b).}
  \label{fig:widthsense}
\end{figure}





\begin{table}[t!]
\centering
\begin{tabular}{@{}cccc@{}}
\toprule
Dataset & $g^*$                                     & Ablated                               & {\textbf{Full}}                         \\ \midrule
ETTh1   & {\ul \textit{1.44  $\pm$ 0.69}} & 1.55 $\pm$ 0.81          & {\textbf{1.24 $\pm$ 0.48}} \\
ETTm1   & 2.12 $\pm$ 7.7            & {\ul \textit{1.30 $\pm$ 0.89}} & {\textbf{1.20 $\pm$ 0.53}} \\
ETTh2   & 1.87 $\pm$ 1.3            & {\ul \textit{1.76 $\pm$ 1.03}} & {\textbf{1.34 $\pm$ 0.62}} \\
ETTm2   & 1.43 $\pm$ 0.93           & {\ul \textit{1.38 $\pm$ 0.63}} & {\textbf{1.19 $\pm$ 0.42}} \\
PEMS08  & 1.45 $\pm$ 0.57           & {\ul \textit{1.35 $\pm$ 0.46}} & {\textbf{1.22 $\pm$ 0.35}} \\
MG      & 1.31 $\pm$ 0.47           & {\ul \textit{1.20 $\pm$ 0.37}} & {\textbf{1.02 $\pm$ 0.09}} \\ \bottomrule
\end{tabular}
\caption{Ablation study restricting the set of candidate models. `Full' has the full set of candidate strategies, `Ablated' cannot select the best fixed strategy. Dynamic strategies, under adversarially selected candidate strategies, can outperform optimal fixed strategies ($g^*$). Tables values use DyStrat-TSF with 10 random seeds.}
\label{Ablation table}
\end{table}

\section{Discussion}
\label{discussion}
The lack of a generally optimal fixed MSF strategy is well documented (\autoref{related}) and burdens practitioners with a search problem over tasks, as well as sub-optimal instance-level performance.
Our proposed method, DyStrat,
addresses both of these issues by automating the search, but at the instance-level such that the resulting forecasts are significantly better performing at both task and instance levels.
Recent studies still only consider finding the best fixed strategy over an entire task \cite{sangiorgio2020robustness_chaotic_multi, chandra2021evaluation_multi_deeplearning, suradhaniwar2021multi, aslam2023multi}. 
Given DyStrat is robust across long and short horizons, and large and small datasets, it shows promise as a simple method for future works to adopt.
\\\newline
The necessity to consider the complexity of hypothesis spaces in all machine learning is well understood \cite{vapnik_overview}. However, time-series forecasting has the unique challenge of specifying the recursive complexity of the data-generating process, represented by the strategy selected. Our work supports a need to further the current theoretical understanding of how the recursive complexity varies within a time-series.
\\\newline
The robustness of DyStrat to sampling candidate strategies, from \autoref{fig:sampled}, opens an interesting discussion on the inter-play between task-level and instance-level ranking. We have shown that including strategies that perform poorly at the task-level, in-fact, have a strong generalisation of good performance at the instance-level. This defines a gap in the literature on how we compare the qualities of forecasting strategies.

\textbf{Limitations and future work} 
The performance of DyStrat is consistent with our hypothesised behaviour. We acknowledge the following as future work to improve the methodology of DyStrat: active learning to better sample training data across forecasters and DMS functions, evaluation with transformers, and strategy ensembling using reinforcement learning \cite{Fu_Wu_Boulet_2022}. 
In applications, we will apply DyStrat to multi-variate time-series tasks.

\textbf{Conclusion.} 
We proposed DyStrat, a method to optimise strategy selection in multi-step forecasting at the instance level.
Our method alleviates the problem of identifying an optimal candidate strategy and generates a dynamic approach to significantly improve forecasting accuracy.
DyStrat consistently outperformed current approaches on long and short term horizons, large and small training data regimes and varying model complexity. We showed robustness to sampling of candidate strategies, presenting an effective addition to any MSF task.

\newpage


\bibliographystyle{unsrtnat}
\bibliography{template}

\newpage
\appendix
\onecolumn

\section{MAE, MAPE, SMAPE, and MAX Metrics}
\label{othermetrics}
We also recorded the Mean absolute error (MAE), mean absolute percentage error (MAPE), Symmetric mean absolute percentage error (SMAPE), and maximum error (MAX). In this appendix section, we only show aggregates of the experiments. We take the ranking of each strategy for each experimental setting; and also record both the overall reduction in error and a number of times the TS-forest method outperformed all fixed strategies.

\begin{figure}[htbp]
    \centering
    \begin{minipage}{0.49\textwidth}
        \centering
        \includegraphics[width=\linewidth]{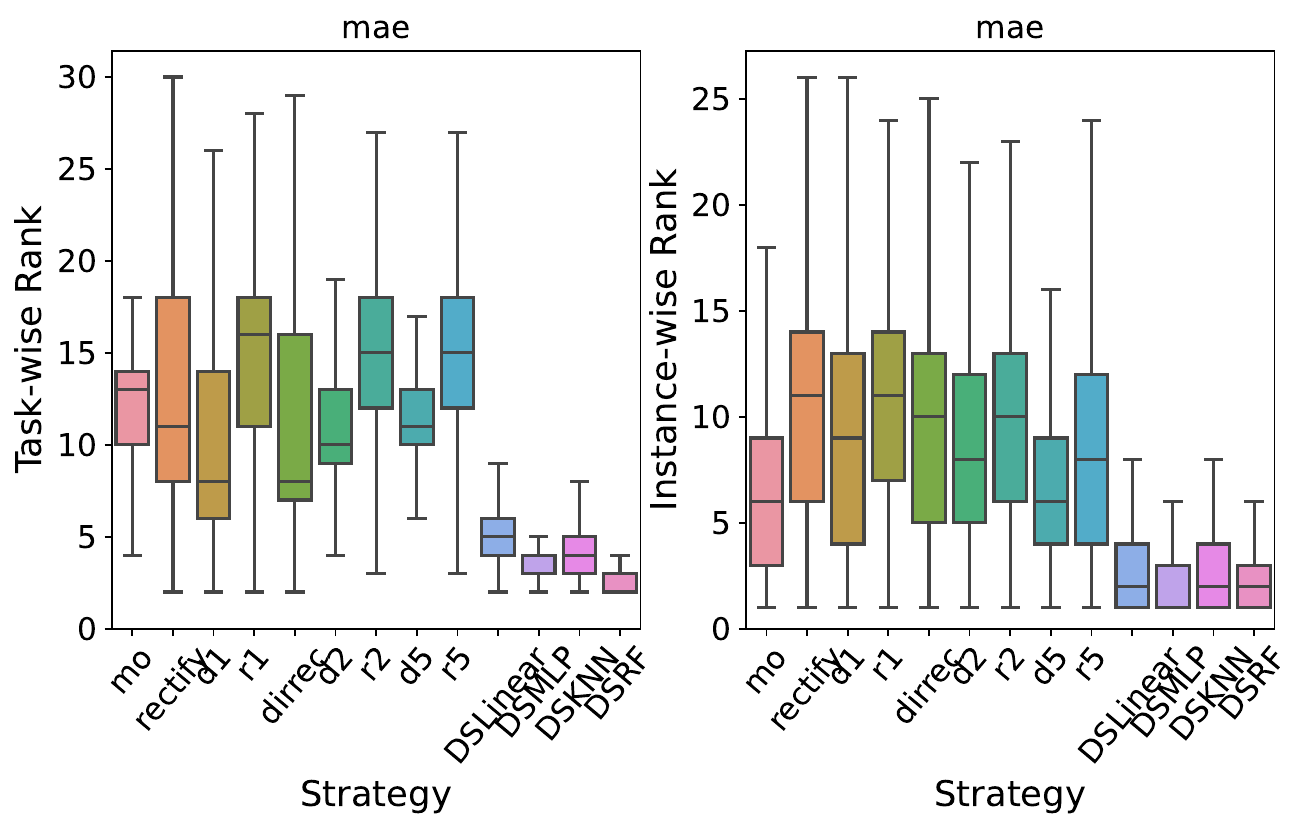}
    \end{minipage}\hfill
    \begin{minipage}{0.49\textwidth}
        \centering
        \includegraphics[width=\linewidth]{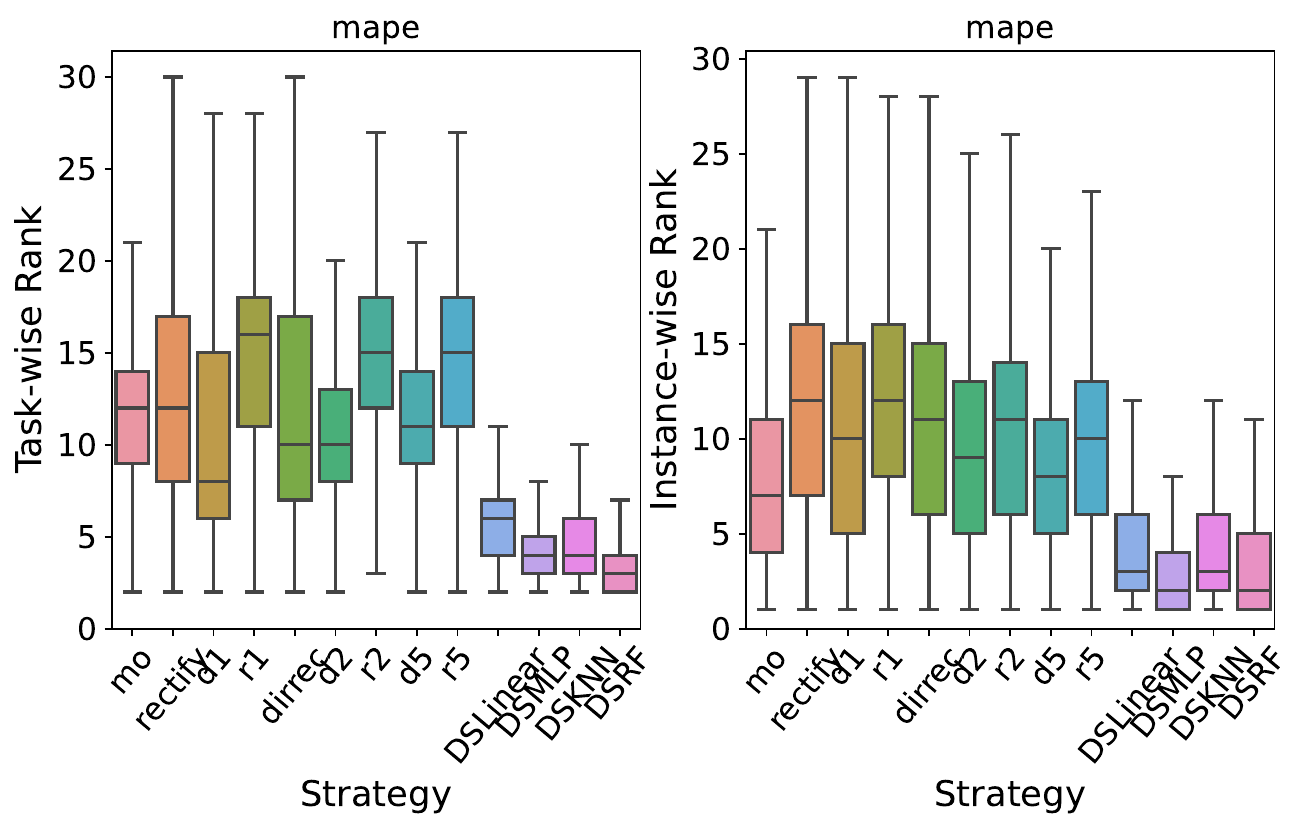}
    \end{minipage}

    \begin{minipage}{0.49\textwidth}
        \centering
        \includegraphics[width=\linewidth]{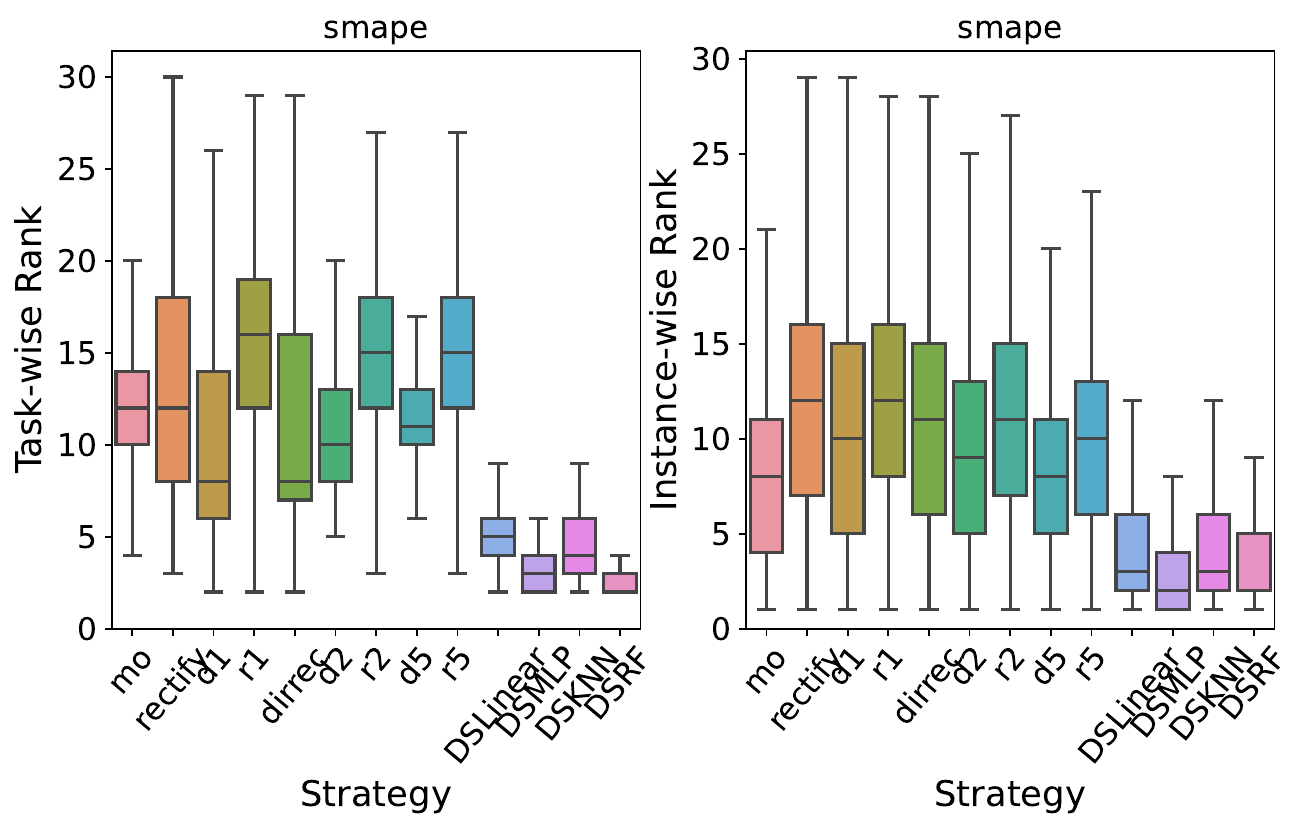}
    \end{minipage}\hfill
    \begin{minipage}{0.49\textwidth}
        \centering
        \includegraphics[width=\linewidth]{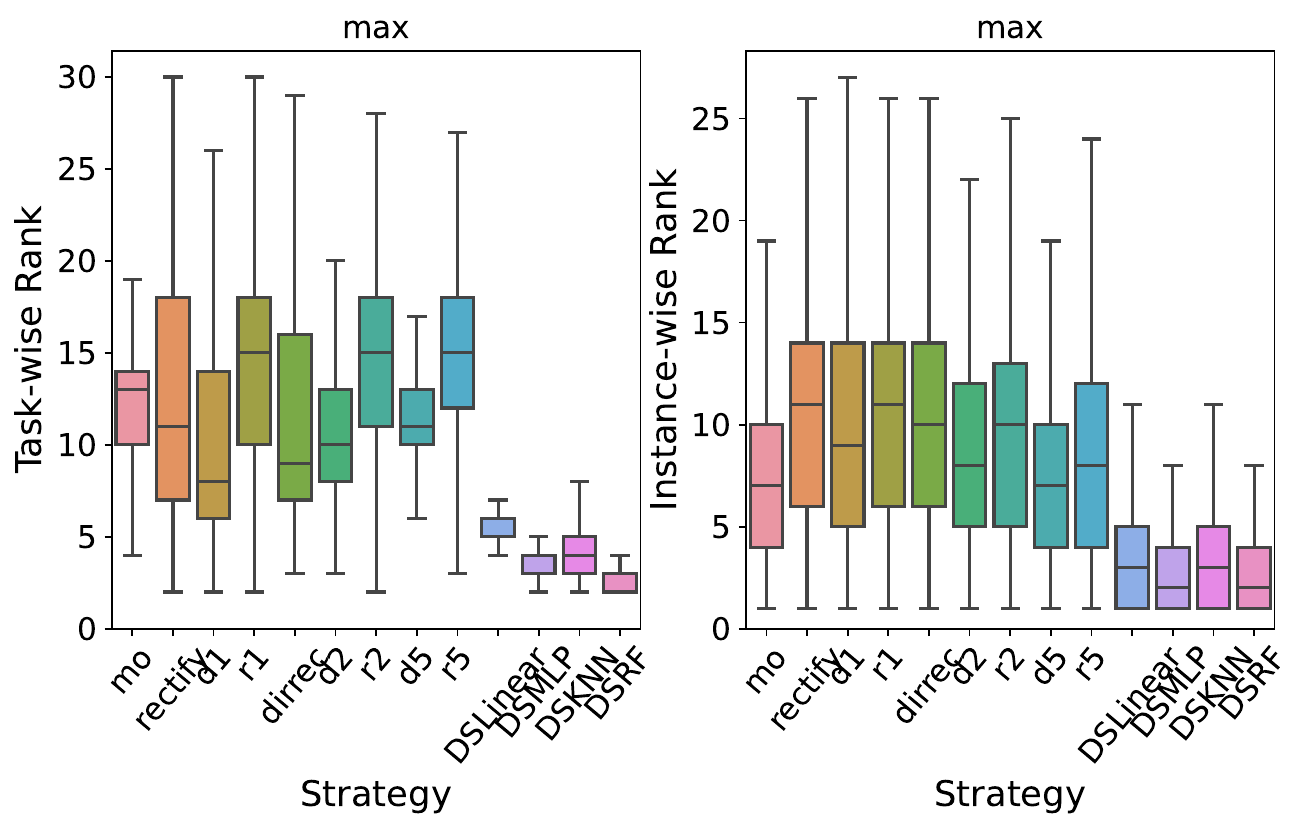}
    \end{minipage}
    \caption{Aggregated results of ranking errors when considering MAE, MAPE, SMAPE, and MAX error-metrics. The metric used is shown at the top of each boxplot figure (top-left MAE, top-right MAPE, bottom-left SMAPE, and bottom-right MAX). Note that DyStrat for these figures was minimising MSE, but outperforms fixed strategies when alternative metrics are used. Left sub-figures show rank over tasks and right sub-figures show ranks over instances.}
\end{figure}

We show that DyStrat, even when the loss function in Algorithm \autoref{zalgo} is minimising MSE, considerably outperforms fixed strategies. When aggregating over tasks, it is notable that the loss function being MSE does not help reduce the MAPE, but \textbf{MSE, MAE, SMAPE, and MAX are all reduced significantly by using DyStrat, in comparison to the best fixed strategy that is not known a priori}.

\begin{table}[hbt!]
\centering
\begin{tabular}{@{}cccccc@{}}
\toprule
Metric    & MSE                             & MAE                             & MAPE                              & SMAPE                           & MAX                             \\ \midrule
$g^*$ \%     & \cellcolor[HTML]{9AFF99}88 - 90 & \cellcolor[HTML]{9AFF99}94 - 96 & \cellcolor[HTML]{FFCCC9}106 - 120 & \cellcolor[HTML]{9AFF99}96 - 98 & \cellcolor[HTML]{9AFF99}93 - 95 \\
\% Better & \cellcolor[HTML]{9AFF99}94      & \cellcolor[HTML]{9AFF99}96      & \cellcolor[HTML]{FFCCC9}63        & \cellcolor[HTML]{9AFF99}89      & \cellcolor[HTML]{9AFF99}88      \\ \bottomrule
\end{tabular}
\caption{$g^*\%$ shows the error of DyStrat using TSForest compared to $g^*$ for that task (a value of 90 means 10\% reduction). \% Better is the proportion of all experiments where DyStrat using TSForest outperforms $g^*$.}
\end{table}

\section{Raw MSE Results}
\label{raw mse results}
Given the vast number of possible metrics for time-series analysis, for space reasons, we only considered the relative error of strategies to that of the optimal dynamic strategy. In this appendix we show the raw MSE and the main results.

\begin{table*}[hbt!]
\centering
\fontsize{8}{12}

\begin{tabular}{@{}lllllllllll@{}}
\toprule
\textit{}          & \textit{ETTh1}    & \textit{ETTh2}    & \textit{ETTm1}    & \textit{ETTm2}    & \textit{MACKY}    & \textit{SINE}     & \textit{LORENZ}   & \textit{TRAFFIC}  & \textit{SUN}      & \textit{POLAND}   \\ \midrule
\textit{mo}        & \textit{3.54e-04} & \textit{3.47e-04} & \textit{5.13e-05} & \textit{6.25e-06} & \textit{1.87e-03} & \textit{1.16e-02} & \textit{7.24e-04} & \textit{5.97e-04} & \textit{4.83e-03} & \textit{3.03e-02} \\
\textit{rectify}   & \textit{5.72e-04} & \textit{1.12e-04} & \textit{9.54e-05} & \textit{1.63e-06} & \textit{9.73e-04} & \textit{3.80e-03} & \textit{2.85e-04} & \textit{5.11e-04} & \textit{3.06e-03} & \textit{8.27e-03} \\
\textit{d1}        & \textit{3.41e-04} & \textit{1.69e-04} & \textit{6.26e-05} & \textit{1.40e-06} & \textit{9.41e-04} & \textit{5.84e-03} & \textit{2.82e-04} & \textit{4.81e-04} & \textit{2.91e-03} & \textit{3.01e-02} \\
\textit{r1}        & \textit{4.24e-04} & \textit{1.53e-04} & \textit{6.52e-05} & \textit{2.81e-06} & \textit{3.23e-02} & \textit{5.03e-03} & \textit{3.50e-03} & \textit{1.10e-03} & \textit{2.84e-03} & \textit{1.48e-02} \\
\textit{dirrec}       & \textit{4.31e-04} & \textit{1.62e-04} & \textit{7.23e-05} & \textit{1.31e-06} & \textit{9.59e-04} & \textit{5.69e-03} & \textit{3.13e-04} & \textit{5.28e-04} & \textit{2.94e-03} & \textit{2.89e-02} \\
\textit{d2}        & \textit{3.67e-04} & \textit{2.05e-04} & \textit{6.35e-05} & \textit{1.72e-06} & \textit{1.16e-03} & \textit{6.68e-03} & \textit{4.24e-04} & \textit{5.05e-04} & \textit{3.23e-03} & \textit{3.74e-02} \\
\textit{r2}        & \textit{4.07e-04} & \textit{2.96e-04} & \textit{6.58e-05} & \textit{3.43e-06} & \textit{2.05e-02} & \textit{8.41e-03} & \textit{7.05e-03} & \textit{1.90e-03} & \textit{5.25e-03} & \textit{8.75e-02} \\
\bottomrule
\end{tabular}
\caption{Raw MSE using 10\% of dataset size for fixed and dynamic strategies. Datasets are shown in column headings and strategies are shown on the left. DS$C$ denotes the dynamic strategy learnt under the learning algorithm $C$.}
\end{table*}

\begin{table*}[hbt!]
\centering
\fontsize{8}{12}
\begin{tabular}{@{}lllllllllll@{}}
\toprule
\textit{}          & \textit{ETTh1}    & \textit{ETTh2}    & \textit{ETTm1}    & \textit{ETTm2}    & \textit{MACKY}    & \textit{SINE}     & \textit{LORENZ}   & \textit{TRAFFIC}  & \textit{SUN}      & \textit{POLAND}   \\ \midrule
\textit{mo}        & \textit{1.84e-04} & \textit{1.54e-04} & \textit{1.73e-05} & \textit{6.86e-07} & \textit{7.20e-04} & \textit{8.60e-03} & \textit{4.20e-04} & \textit{3.74e-04} & \textit{3.42e-03} & \textit{1.85e-02} \\
\textit{rectify}   & \textit{3.08e-04} & \textit{5.19e-05} & \textit{5.41e-05} & \textit{7.52e-07} & \textit{3.45e-04} & \textit{2.04e-03} & \textit{9.22e-05} & \textit{2.70e-04} & \textit{2.25e-03} & \textit{7.77e-03} \\
\textit{d1}        & \textit{1.90e-04} & \textit{5.52e-05} & \textit{3.06e-05} & \textit{3.96e-07} & \textit{2.79e-04} & \textit{2.06e-03} & \textit{1.06e-04} & \textit{3.19e-04} & \textit{2.25e-03} & \textit{2.32e-02} \\
\textit{r1}        & \textit{2.17e-04} & \textit{6.93e-05} & \textit{3.01e-05} & \textit{1.90e-06} & \textit{2.22e-02} & \textit{1.70e-03} & \textit{4.87e-03} & \textit{9.80e-04} & \textit{2.52e-03} & \textit{1.20e-02} \\
\textit{dirrec}       & \textit{2.02e-04} & \textit{5.90e-05} & \textit{3.53e-05} & \textit{4.35e-07} & \textit{2.87e-04} & \textit{1.96e-03} & \textit{9.30e-05} & \textit{3.05e-04} & \textit{2.32e-03} & \textit{1.64e-02} \\
\textit{d2}        & \textit{1.93e-04} & \textit{7.19e-05} & \textit{2.85e-05} & \textit{3.69e-07} & \textit{3.67e-04} & \textit{2.69e-03} & \textit{1.34e-04} & \textit{3.25e-04} & \textit{2.62e-03} & \textit{2.95e-02} \\
\textit{r2}        & \textit{2.13e-04} & \textit{1.15e-04} & \textit{3.08e-05} & \textit{6.44e-07} & \textit{1.63e-02} & \textit{3.26e-03} & \textit{3.10e-03} & \textit{1.03e-03} & \textit{3.36e-03} & \textit{3.95e-02} \\
\bottomrule
\end{tabular}
\caption{Raw MSE using 20\% of dataset size for fixed and dynamic strategies. Datasets are shown in column headings and strategies are shown on the left. DS$C$ denotes the dynamic strategy learnt under the learning algorithm $C$.}\end{table*}

\begin{table*}[hbt!]
\centering
\fontsize{8}{12}
\begin{tabular}{@{}lllllllllll@{}}
\toprule
\textit{}          & \textit{ETTh1}    & \textit{ETTh2}    & \textit{ETTm1}    & \textit{ETTm2}    & \textit{MACKY}    & \textit{SINE}     & \textit{LORENZ}   & \textit{TRAFFIC}  & \textit{SUN}      & \textit{POLAND}   \\ \midrule
\textit{mo}        & \textit{6.89e-05} & \textit{3.21e-05} & \textit{5.16e-06} & \textit{3.26e-07} & \textit{3.88e-04} & \textit{2.44e-03} & \textit{1.07e-04} & \textit{2.57e-04} & \textit{2.60e-03} & \textit{5.00e-03} \\
\textit{rectify}   & \textit{1.43e-04} & \textit{2.44e-05} & \textit{1.74e-05} & \textit{4.76e-07} & \textit{1.59e-04} & \textit{1.01e-03} & \textit{2.12e-05} & \textit{2.21e-04} & \textit{8.70e-04} & \textit{2.21e-03} \\
\textit{d1}        & \textit{9.50e-05} & \textit{1.96e-05} & \textit{1.15e-05} & \textit{1.79e-07} & \textit{1.55e-04} & \textit{6.16e-04} & \textit{2.45e-05} & \textit{1.83e-04} & \textit{1.07e-03} & \textit{2.54e-03} \\
\textit{r1}        & \textit{1.02e-04} & \textit{3.08e-05} & \textit{1.69e-05} & \textit{3.58e-07} & \textit{1.40e-02} & \textit{9.11e-04} & \textit{3.31e-03} & \textit{7.88e-04} & \textit{9.49e-04} & \textit{2.18e-03} \\
\textit{dirrec}       & \textit{1.10e-04} & \textit{2.04e-05} & \textit{1.32e-05} & \textit{2.53e-07} & \textit{1.42e-04} & \textit{6.17e-04} & \textit{3.20e-05} & \textit{2.05e-04} & \textit{1.07e-03} & \textit{2.22e-03} \\
\textit{d2}        & \textit{9.37e-05} & \textit{2.14e-05} & \textit{9.01e-06} & \textit{1.85e-07} & \textit{1.86e-04} & \textit{9.01e-04} & \textit{4.49e-05} & \textit{2.10e-04} & \textit{1.54e-03} & \textit{2.74e-03} \\
\textit{r2}        & \textit{1.08e-04} & \textit{3.35e-05} & \textit{1.37e-05} & \textit{1.31e-07} & \textit{1.67e-02} & \textit{1.11e-03} & \textit{2.92e-03} & \textit{1.02e-03} & \textit{1.81e-03} & \textit{3.63e-03} \\
\bottomrule
\end{tabular}
\caption{Raw MSE using 40\% of dataset size for fixed and dynamic strategies. Datasets are shown in column headings and strategies are shown on the left. DS$C$ denotes the dynamic strategy learnt under the learning algorithm $C$.}\end{table*}

\begin{table*}[hbt!]
\centering
\fontsize{8}{12}
\begin{tabular}{@{}lllllllllll@{}}
\toprule
\textit{}                     & \textit{ETTh1}                   & \textit{ETTh2}                   & \textit{ETTm1}                   & \textit{ETTm2}                   & \textit{MACKY}                   & \textit{SINE}                    & \textit{LORENZ}                  & \textit{TRAFFIC}                 & \textit{SUN}                     & \textit{POLAND}                  \\ \midrule
\textit{mo}                   & \textit{2.25e-05}                & \textit{8.57e-06}                & \textit{1.60e-06}                & \textit{2.09e-07}                & \textit{2.02e-04}                & \textit{6.44e-04}                & \textit{3.57e-05}                & \textit{1.47e-04}                & \textit{1.19e-03}                & \textit{1.56e-03}                \\
\textit{rectify}              & \textit{8.64e-05}                & \textit{1.45e-05}                & \textit{4.32e-06}                & \textit{5.77e-07}                & \textit{8.92e-05}                & \textit{5.25e-04}                & \textit{1.20e-05}                & \textit{1.06e-04}                & \textit{2.90e-04}                & \textit{9.47e-04}                \\
\textit{d1}                   & \textit{4.62e-05}                & \textit{1.01e-05}                & \textit{3.01e-06}                & \textit{7.58e-07}                & \textit{6.76e-05}                & \textit{2.76e-04}                & \textit{1.06e-05}                & \textit{1.05e-04}                & \textit{3.17e-04}                & \textit{7.29e-04}                \\
\textit{r1}                   & \textit{4.78e-05}                & \textit{1.55e-05}                & \textit{4.57e-06}                & \textit{2.16e-06}                & \textit{1.77e-02}                & \textit{3.43e-04}                & \textit{2.64e-03}                & \textit{5.09e-04}                & \textit{3.12e-04}                & \textit{1.66e-03}                \\
\bottomrule
\end{tabular}
\caption{Raw MSE using 80\% of dataset size for fixed and dynamic strategies. Datasets are shown in column headings and strategies are shown on the left. DS$C$ denotes the dynamic strategy learnt under the learning algorithm $C$.}\end{table*}

Across all sizes of training data percentage, the MSE of dynamic strategies under the TSForest were the lowest.

\begin{table*}[hbt!]
\centering
\fontsize{8}{12}
\begin{tabular}{@{}lllllllll@{}}
\toprule
\textit{}                     & \textit{ETTh1}                   & \textit{ETTh2}                   & \textit{ETTm1}                   & \textit{ETTm2}                   & \textit{MACKY}                   & \textit{SINE}                    & \textit{LORENZ}                  & \textit{TRAFFIC}                 \\ \midrule
\textit{mo}                   & \textit{1.75e-05}                & \textit{4.94e-06}                & \textit{3.31e-06}                & \textit{2.23e-07}                & \textit{1.79e-04}                & \textit{4.62e-04}                & \textit{2.14e-05}                & \textit{1.28e-04}                \\
\textit{rectify}              & \textit{3.33e-05}                & \textit{8.68e-06}                & \textit{3.16e-06}                & \textit{3.28e-07}                & \textit{7.43e-05}                & \textit{2.44e-04}                & \textit{6.95e-06}                & \textit{6.05e-05}                \\
\textit{d1}                   & \textit{2.35e-05}                & \textit{4.72e-06}                & \textit{2.18e-06}                & \textit{4.20e-07}                & \textit{7.60e-05}                & \textit{1.63e-04}                & \textit{9.32e-06}                & \textit{7.95e-05}                \\
\textit{r1}                   & \textit{2.27e-05}                & \textit{1.28e-05}                & \textit{2.24e-06}                & \textit{2.84e-07}                & \textit{9.18e-03}                & \textit{1.74e-04}                & \textit{1.24e-03}                & \textit{2.84e-04}                \\
\textit{dirrec}                  & \textit{2.88e-05}                & \textit{5.81e-06}                & \textit{3.65e-06}                & \textit{6.38e-07}                & \textit{7.81e-05}                & \textit{1.67e-04}                & \textit{8.50e-06}                & \textit{7.83e-05}                \\
\bottomrule
\end{tabular}
\caption{Raw MSE for horizon length 10 for fixed and dynamic strategies. Datasets are shown in column headings and strategies are shown on the left. DS$C$ denotes the dynamic strategy learnt under the learning algorithm $C$.}\end{table*}

\begin{table*}[hbt!]
\centering
\fontsize{8}{12}

\begin{tabular}{@{}lllllllll@{}}
\toprule
\textit{}                     & \textit{ETTh1}                   & \textit{ETTh2}                   & \textit{ETTm1}                   & \textit{ETTm2}                   & \textit{MACKY}                   & \textit{SINE}                    & \textit{LORENZ}                  & \textit{TRAFFIC}                 \\ \midrule
\textit{mo}                   & \textit{2.40e-05}                & \textit{9.02e-06}                & \textit{1.65e-06}                & \textit{2.42e-07}                & \textit{2.27e-04}                & \textit{7.99e-04}                & \textit{4.08e-05}                & \textit{1.62e-04}                \\
\textit{rectify}              & \textit{7.20e-05}                & \textit{2.05e-05}                & \textit{5.72e-06}                & \textit{5.97e-07}                & \textit{9.52e-05}                & \textit{5.43e-04}                & \textit{1.37e-05}                & \textit{1.08e-04}                \\
\textit{d1}                   & \textit{5.27e-05}                & \textit{1.00e-05}                & \textit{3.79e-06}                & \textit{6.36e-07}                & \textit{7.34e-05}                & \textit{3.04e-04}                & \textit{1.43e-05}                & \textit{1.07e-04}                \\
\textit{r1}                   & \textit{7.70e-05}                & \textit{1.40e-05}                & \textit{4.88e-06}                & \textit{5.39e-07}                & \textit{1.94e-02}                & \textit{3.88e-04}                & \textit{2.96e-03}                & \textit{7.03e-04}                \\
\textit{dirrec}                  & \textit{5.51e-05}                & \textit{1.26e-05}                & \textit{4.08e-06}                & \textit{9.54e-07}                & \textit{7.86e-05}                & \textit{3.33e-04}                & \textit{1.14e-05}                & \textit{1.12e-04}                \\
\bottomrule
\end{tabular}
\caption{Raw MSE for horizon length 20 for fixed and dynamic strategies. Datasets are shown in column headings and strategies are shown on the left. DS$C$ denotes the dynamic strategy learnt under the learning algorithm $C$.}
\end{table*}

\begin{table}[hbt!]
\centering
\fontsize{8}{12}

\begin{tabular}{@{}lllllllll@{}}
\toprule
\textit{}                     & \textit{ETTh1}                   & \textit{ETTh2}                   & \textit{ETTm1}                   & \textit{ETTm2}                   & \textit{MACKY}                   & \textit{SINE}                    & \textit{LORENZ}                  & \textit{TRAFFIC}                 \\ \midrule
\textit{mo}                   & \textit{3.28e-05}                & \textit{1.60e-05}                & \textit{1.79e-06}                & \textit{3.23e-07}                & \textit{2.65e-04}                & \textit{1.41e-03}                & \textit{7.78e-05}                & \textit{1.99e-04}                \\
\textit{rectify}              & \textit{1.40e-04}                & \textit{4.45e-05}                & \textit{8.02e-06}                & \textit{1.17e-06}                & \textit{1.29e-04}                & \textit{1.09e-03}                & \textit{3.80e-05}                & \textit{1.81e-04}                \\
\textit{d1}                   & \textit{9.00e-05}                & \textit{1.97e-05}                & \textit{4.79e-06}                & \textit{1.11e-06}                & \textit{8.30e-05}                & \textit{6.19e-04}                & \textit{2.53e-05}                & \textit{1.59e-04}                \\
\textit{r1}                   & \textit{6.85e-05}                & \textit{6.40e-05}                & \textit{7.15e-06}                & \textit{1.19e-06}                & \textit{1.52e-02}                & \textit{3.63e-03}                & \textit{4.78e-03}                & \textit{1.22e-03}                \\
\textit{dirrec}                  & \textit{1.13e-04}                & \textit{2.35e-05}                & \textit{5.37e-06}                & \textit{1.44e-06}                & \textit{8.96e-05}                & \textit{6.42e-04}                & \textit{2.65e-05}                & \textit{1.66e-04}                \\
\bottomrule
\end{tabular}

\caption{Raw MSE for horizon length 40 for fixed and dynamic strategies. Datasets are shown in column headings and strategies are shown on the left. DS$C$ denotes the dynamic strategy learnt under the learning algorithm $C$.}
\end{table}

\begin{table}[hbt!]
\centering
\fontsize{8}{12}

\begin{tabular}{@{}lllllllll@{}}
\toprule
\textit{}                     & \textit{ETTh1}                   & \textit{ETTh2}                   & \textit{ETTm1}                   & \textit{ETTm2}                   & \textit{MACKY}                   & \textit{SINE}                    & \textit{LORENZ}                  & \textit{TRAFFIC}                 \\ \midrule
\textit{mo}                   & \textit{3.92e-05}                & \textit{2.61e-05}                & \textit{2.06e-06}                & \textit{5.25e-07}                & \textit{3.68e-04}                & \textit{2.17e-03}                & \textit{1.39e-04}                & \textit{3.44e-04}                \\
\textit{rectify}              & \textit{4.31e-04}                & \textit{8.86e-05}                & \textit{6.99e-06}                & \textit{1.72e-06}                & \textit{2.73e-04}                & \textit{1.68e-03}                & \textit{8.59e-05}                & \textit{4.02e-04}                \\
\textit{d1}                   & \textit{1.91e-04}                & \textit{4.13e-05}                & \textit{5.28e-06}                & \textit{1.95e-06}                & \textit{1.45e-04}                & \textit{1.04e-03}                & \textit{6.07e-05}                & \textit{2.65e-04}                \\
\textit{r1}                   & \textit{2.77e-04}                & \textit{5.47e-05}                & \textit{5.92e-06}                & \textit{1.88e-06}                & \textit{3.17e-02}                & \textit{1.40e-02}                & \textit{8.10e-04}                & \textit{2.69e-03}                \\
\textit{dirrec}                  & \textit{2.32e-04}                & \textit{4.94e-05}                & \textit{6.22e-06}                & \textit{2.55e-06}                & \textit{1.56e-04}                & \textit{1.14e-03}                & \textit{6.39e-05}                & \textit{2.86e-04}                \\
\bottomrule
\end{tabular}
\caption{Raw MSE for horizon length 80 for fixed and dynamic strategies. Datasets are shown in column headings and strategies are shown on the left. DS$C$ denotes the dynamic strategy learnt under the learning algorithm $C$.}
\end{table}

\begin{table}[hbt!]
\centering
\fontsize{8}{12}

\begin{tabular}{@{}lllllllll@{}}
\toprule
\textit{}                     & \textit{ETTh1}                   & \textit{ETTh2}                   & \textit{ETTm1}                   & \textit{ETTm2}                   & \textit{MACKY}                   & \textit{SINE}                    & \textit{LORENZ}                  & \textit{TRAFFIC}                 \\ \midrule
\textit{mo}                   & \textit{9.35e-05}                & \textit{5.19e-05}                & \textit{3.06e-06}                & \textit{1.20e-06}                & \textit{1.07e-03}                & \textit{2.73e-03}                & \textit{2.76e-04}                & \textit{4.20e-04}                \\
\textit{rectify}              & \textit{1.08e-03}                & \textit{1.81e-04}                & \textit{1.48e-05}                & \textit{2.54e-06}                & \textit{1.19e-03}                & \textit{2.78e-03}                & \textit{2.34e-04}                & \textit{5.02e-04}                \\
\textit{d1}                   & \textit{6.06e-04}                & \textit{8.23e-05}                & \textit{1.03e-05}                & \textit{3.18e-06}                & \textit{5.98e-04}                & \textit{1.78e-03}                & \textit{1.67e-04}                & \textit{3.56e-04}                \\
\textit{r1}                   & \textit{7.69e-04}                & \textit{1.38e-04}                & \textit{1.21e-05}                & \textit{4.79e-06}                & \textit{1.23e-02}                & \textit{4.25e-02}                & \textit{9.39e-04}                & \textit{4.24e-03}                \\
\textit{dirrec}                  & \textit{7.29e-04}                & \textit{9.54e-05}                & \textit{1.37e-05}                & \textit{4.17e-06}                & \textit{6.52e-04}                & \textit{2.23e-03}                & \textit{1.71e-04}                & \textit{3.88e-04}                \\
\bottomrule
\end{tabular}
\caption{Raw MSE for horizon length 160 for fixed and dynamic strategies. Datasets are shown in column headings and strategies are shown on the left. DS$C$ denotes the dynamic strategy learnt under the learning algorithm $C$.}
\end{table}

Across all lengths of multi-step horizon, the MSE of dynamic strategies under the TSForest were the lowest.






\end{document}